\documentclass{article}
\usepackage[final]{neurips_2023}

\PassOptionsToPackage{dvipsnames}{xcolor}
\usepackage{microtype}
\usepackage{graphicx}
\usepackage{booktabs} %

\usepackage{hyperref}

\usepackage{amsmath}
\usepackage{amssymb}
\usepackage{mathtools}
\usepackage{amsthm}

\usepackage[capitalize,noabbrev]{cleveref}

\usepackage[utf8]{inputenc} %
\usepackage[T1]{fontenc}    %
\usepackage{hyperref}       %
\usepackage{url}            %
\usepackage{nicefrac}       %
\usepackage{xcolor}
\usepackage{xspace}

\usepackage{amsfonts, dsfont} %
\usepackage{amsmath}
\usepackage{amsthm}
\usepackage{amssymb}
\usepackage{xspace}  %
\usepackage{stmaryrd}
\usepackage{url}
\usepackage{enumitem}
\usepackage{wrapfig}
\usepackage{caption}
\usepackage{algorithm}
\usepackage{algorithmicx}
\usepackage[noend]{algpseudocode}
\usepackage{algpseudocode}
\DeclareCaptionSubType*{algorithm}
\definecolor{ourspecialtextcolor}{rgb}{0.528, 0.471, 0.701}
\renewcommand{\algorithmiccomment}[1]{\bgroup\hskip1em\textcolor{ourspecialtextcolor}{//~\textsl{#1}}\egroup}

\usepackage{mathtools}
\usepackage{adjustbox}
\usepackage{subcaption}
\usepackage{bm}
\usepackage{manfnt}

\usepackage{chngcntr}

\newcommand{\vv}[1]{\boldsymbol{#1}}
\newcommand{\id}[1]{\llbracket{#1}\rrbracket}

\newcommand{\probsmt}{\ensuremath{\mathcal{M}}\xspace} %

\newcommand{\cvars}{\ensuremath{\mathbf{X}}\xspace} %
\DeclareMathOperator{\E}{\mathds{E}}
\newcommand{\R}{\ensuremath{\mathbb{R}}\xspace} %

\newcommand{\data}{\ensuremath{\mathcal{D}}\xspace} %
\newcommand{\literals}{\ensuremath{\mathcal{L}}\xspace} %
\newcommand{\literal}{\ensuremath{\ell}\xspace} %
\newcommand{\Weights}{\ensuremath{\Phi}\xspace} %
\newcommand{\weight}{\ensuremath{\phi}\xspace} %

\newcommand{\x}{\ensuremath{\vv{x}}\xspace}
\newcommand{\y}{\ensuremath{y}\xspace}

\newcommand{\theory}{\ensuremath{\Delta}}

\newcommand{\smt}{SMT\xspace}

\newcommand{\WMI}{\ensuremath{\mathsf{WMI}}\xspace}
\newcommand{\X}{\ensuremath{\bm{X}}}

\newcommand{\true}[0]{\texttt{true}\xspace}

\newcommand{\relu}{\ensuremath{\mathsf{ReLU}}\xspace}

\newcommand{\prediction}{\ensuremath{\mathit{pred}}\xspace}
\newcommand{\posterior}{\ensuremath{\mathit{pos}}\xspace}

\newcommand{\nn}{\ensuremath{f}\xspace}
\newcommand{\fw}{\ensuremath{\bm{w}}\xspace}

\newcommand{\gaussian}{\ensuremath{\mathcal{N}}\xspace}
\newcommand{\gaussianp}{\ensuremath{p_{\gaussian}}\xspace}

\newcommand{\allrelu}{\ensuremath{\mathcal{R}}\xspace}
\newcommand{\reluinput}{\ensuremath{r}\xspace}
\newcommand{\bidx}{\ensuremath{\bm{B}}\xspace}

\newcommand{\sampling}{\ensuremath{s}\xspace}
\newcommand{\collapse}{\ensuremath{c}\xspace}

\newcommand{\W}{\ensuremath{\bm{W}}\xspace}
\newcommand{\sampleW}{\ensuremath{\mathcal{W}}\xspace}
\newcommand{\variance}{\ensuremath{\sigma^2}\xspace}
\newcommand{\Y}{\ensuremath{\bm{Y}}\xspace}
\newcommand{\ciber}{CIBER\xspace}

\newcommand{\predictivep}{\ensuremath{p(\y \mid \x, \fw)}\xspace}

\newcommand{\polytope}{\ensuremath{\mbox{\mancube}}\xspace}
\newcommand{\wvcf}{\ensuremath{\phi}\xspace}

\newcommand{\radius}{\ensuremath{r}\xspace}
\newcommand{\trir}{\ensuremath{\alpha}\xspace}

\DeclareMathOperator*{\argmax}{arg\,max}

\newenvironment{shrinkeq}[1]
{\bgroup
\addtolength\abovedisplayshortskip{#1}
\addtolength\abovedisplayskip{#1}
\addtolength\belowdisplayshortskip{#1}
\addtolength\belowdisplayskip{#1}}
{\egroup\ignorespacesafterend}

\newtheorem{thm}{Theorem}

\newtheorem{mydef}[thm]{Definition}

\newtheorem{exa}[thm]{Example} 
 
\newtheorem{pro}[thm]{Proposition}

\theoremstyle{plain}

\theoremstyle{definition}

\theoremstyle{remark}

\title{Collapsed Inference for Bayesian Deep Learning}

\author{
Zhe Zeng \\
 Computer Science Department\\
University of California, Los Angeles \\
\texttt{zhezeng@cs.ucla.edu} \\
\And
Guy Van den Broeck \\
Computer Science Department\\
University of California, Los Angeles \\
\texttt{guyvdb@cs.ucla.edu} \\
}

\begin{document}

\maketitle

\begin{abstract}
Bayesian neural networks~(BNNs) provide a formalism to quantify and calibrate uncertainty in deep learning.
Current inference approaches for BNNs often resort to few-sample estimation for scalability,
which can harm predictive performance,
while its alternatives tend to be computationally prohibitively expensive.
We tackle this challenge by revealing a previously unseen connection between inference on BNNs and \emph{volume computation} problems.
With this observation, we introduce a novel collapsed inference scheme that performs Bayesian model averaging using \emph{collapsed samples}.
It improves over a Monte-Carlo sample by limiting sampling to a subset of the network weights while pairing it with some closed-form conditional distribution over the rest.
A collapsed sample represents uncountably many models drawn from the approximate posterior and thus yields higher sample efficiency.
Further, we show that the marginalization of a collapsed sample can be solved analytically and efficiently despite the non-linearity of neural networks by leveraging existing volume computation solvers.
Our proposed use of collapsed samples achieves a balance between scalability and accuracy. 
On various regression and classification tasks, our collapsed Bayesian deep learning approach demonstrates significant improvements over existing methods and sets a new state of the art in terms of uncertainty estimation as well as predictive performance.

\end{abstract}

\section{Introduction}
\begin{figure}[t]
\centering
\hfill
\begin{subfigure}{0.45\textwidth}
  \centering
  \includegraphics[width=.70\textwidth]{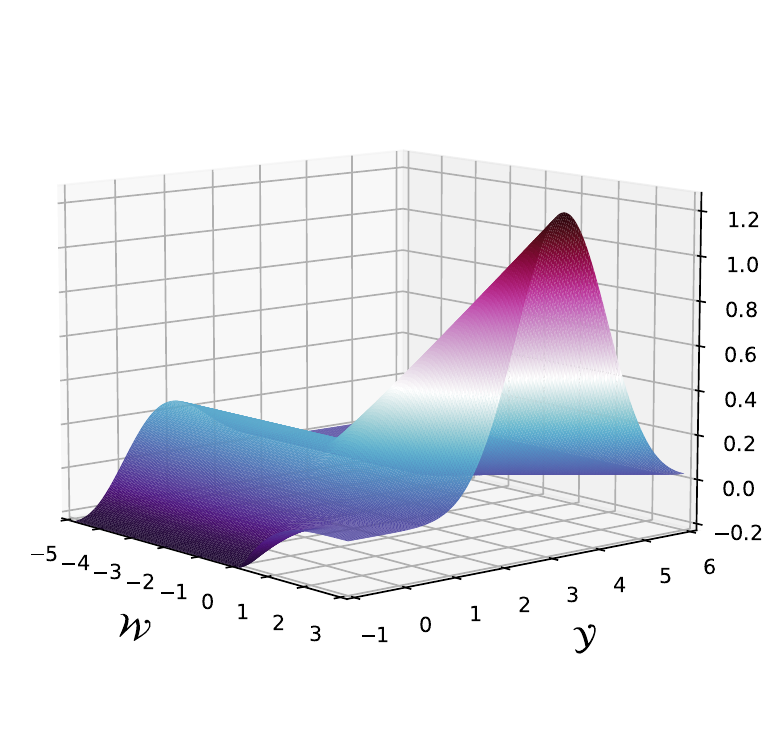}
  \caption{$p(\y \mid \x, \fw)$ being Gaussian. See Example~\ref{exa:gaussian avg relu}.}
  \label{fig:gaussian surface}
\end{subfigure}
\hfill
\begin{subfigure}{0.45\textwidth}
  \centering
\includegraphics[width=.70\textwidth]{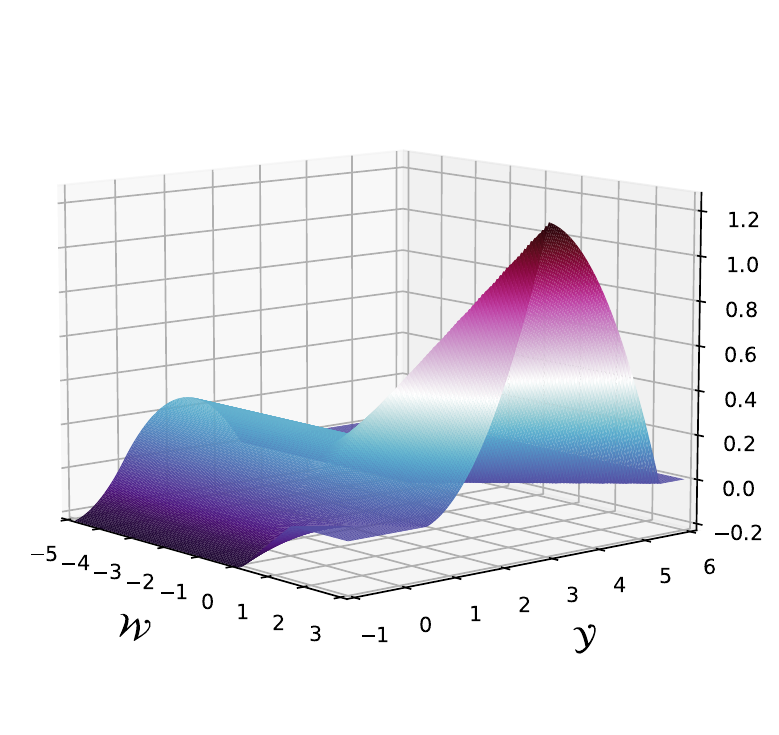}
  \caption{$p(\y \mid \x, \fw)$ being triangular. See Section~\ref{sec: exact BMA over relu}.}
  \label{fig:triangular surface}
\end{subfigure}
\hfill
\caption{
The integral surface of (a) the expected prediction in BMA, and 
(b) our proposed approximation. Both are highly non-convex and multi-modal.
The z-axis is the weighted prediction $\y~p(\y \mid \x, \fw)~p(\fw \mid \data)$. 
Integration of~(a) does not admit a closed-form solution, yet integration of~(b) is a close approximation that can be solved exactly and efficiently by WMI solvers.
}
\label{fig:test}
\end{figure}

Uncertainty estimation is crucial for decision making.
Deep learning models, including those in safety-critical domains, tend to estimate uncertainty poorly.
To overcome this issue, Bayesian deep learning obtains a posterior distribution over the model parameters 
hoping to improve predictions and provide reliable uncertainty estimates.
Among Bayesian inference procedures with neural networks, 
Bayesian model averaging~(BMA) is particularly compelling~\citep{wasserman2000bayesian,fragoso2018bayesian,maddox2019simple}.
However, computing BMAs is distinctly challenging since it involves marginalizing over posterior parameters,
which possess some unusual topological properties such as mode-connectivity \citep{izmailov2021bayesian}.
We show that even with simple low-dimensional approximate parameter posteriors as uniform distributions,
doing BMA requires integrating over highly \emph{non-convex} and \emph{multi-modal} distributions with discontinuities arising from non-linear activations (cf.~Figure~\ref{fig:gaussian surface}).
Accurately approximating the BMA can achieve significant performance gains \citep{izmailov2021bayesian}.
Existing methods mainly focus on general-purpose MCMC, which can fail to converge,
or provides inaccurate few-sample predictions~\citep{kristiadiposterior},
because running longer sampling chains is computationally expensive,
and variational approaches that typically use 
a mean-field approximation that ignores correlations induced by activations~\citep{jospin2022hands}.

In this work, we are interested in developing \emph{collapsed samplers}, also known as \emph{cutset} or \emph{Rao-Blackwellised} samplers for BMA.
A collapsed sampler improves over classical particle-based methods by limiting sampling to a subset of variables and further pairing each sample with a closed-form representation of a conditional distribution over the rest whose marginalization is often tractable.
Collapsed samplers are effective at variance reduction in graphical models~\citep{koller2009probabilistic}, however
no collapsed samplers are known for Bayesian deep learning.
We believe that this is due to the lack of a closed-form marginalization technique congruous with the non-linearity in deep neural networks.
Our aim is to overcome this issue and improve BMA estimation by 
incorporating exact marginalization over (close approximate) conditional distributions into the inference scheme.
Nevertheless, scalability and efficiency are guaranteed by the sampling part of our proposed algorithm.

\begin{figure*}[t]
    \centering
    \begin{subfigure}{0.31\textwidth}
    \centering
    \includegraphics[width=0.99\textwidth]{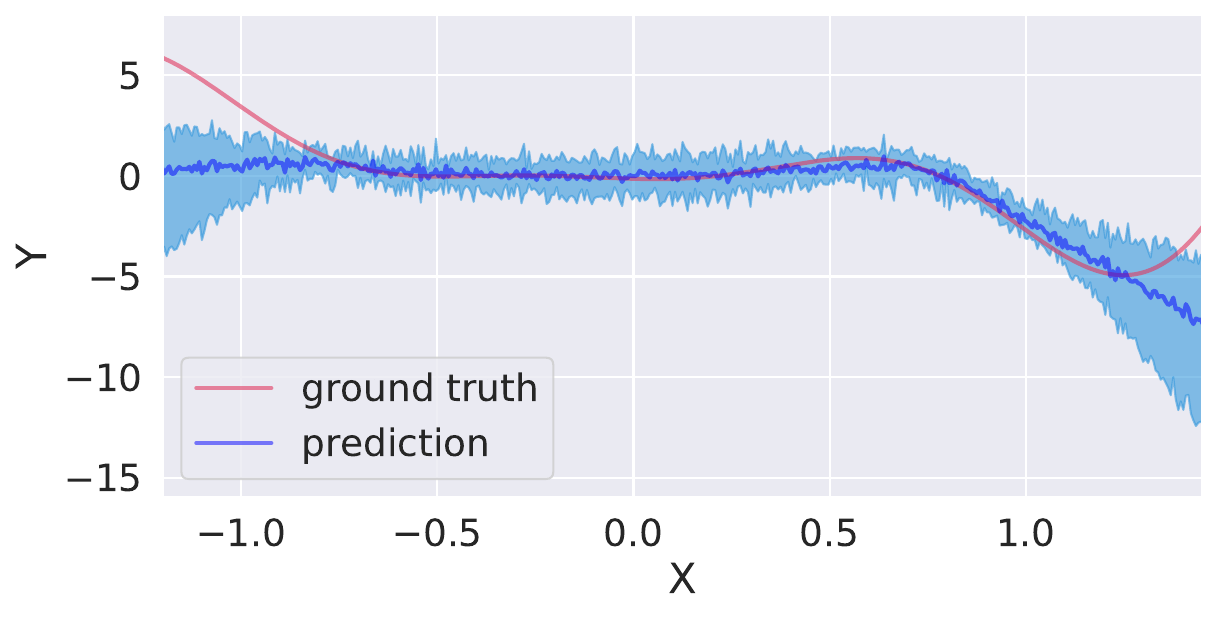}
    \caption{HMC with $10$ samples}
    \label{fig:toy hmc few}
    \end{subfigure}
    \hfill
    \begin{subfigure}{0.34\textwidth}
    \centering
    \includegraphics[width=0.903\textwidth]{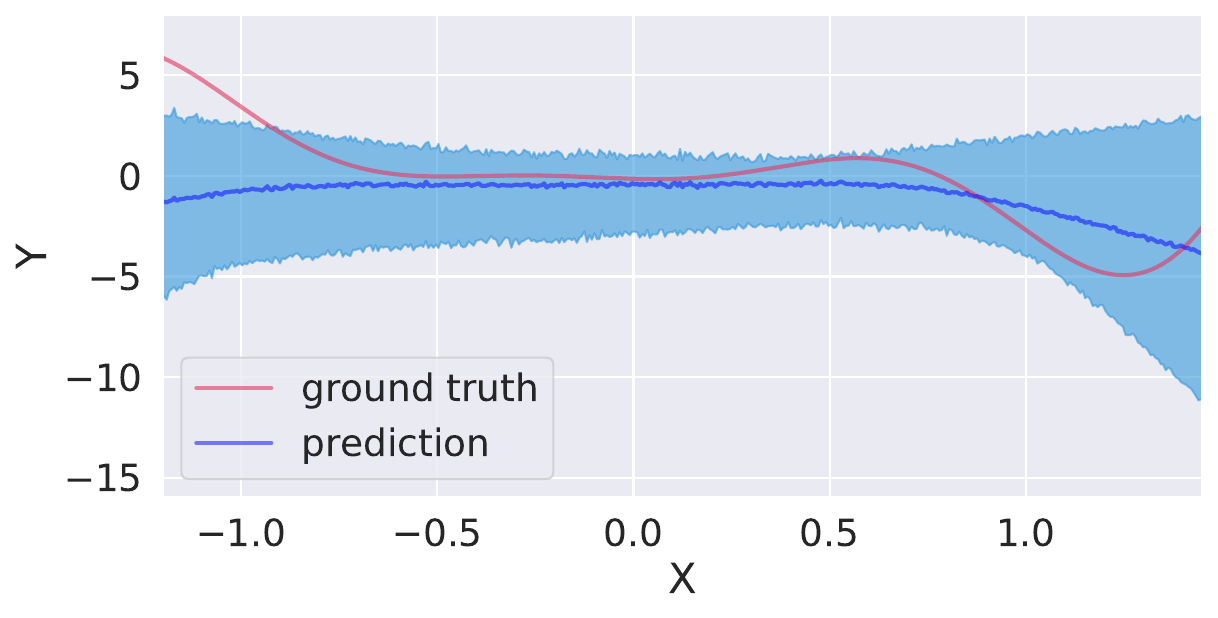}
    \caption{CIBER with $10$ collapsed samples}
    \label{fig:toy ciber}
    \end{subfigure}
    \hfill
    \begin{subfigure}{0.31\textwidth}
    \centering
    \includegraphics[width=0.99\textwidth]{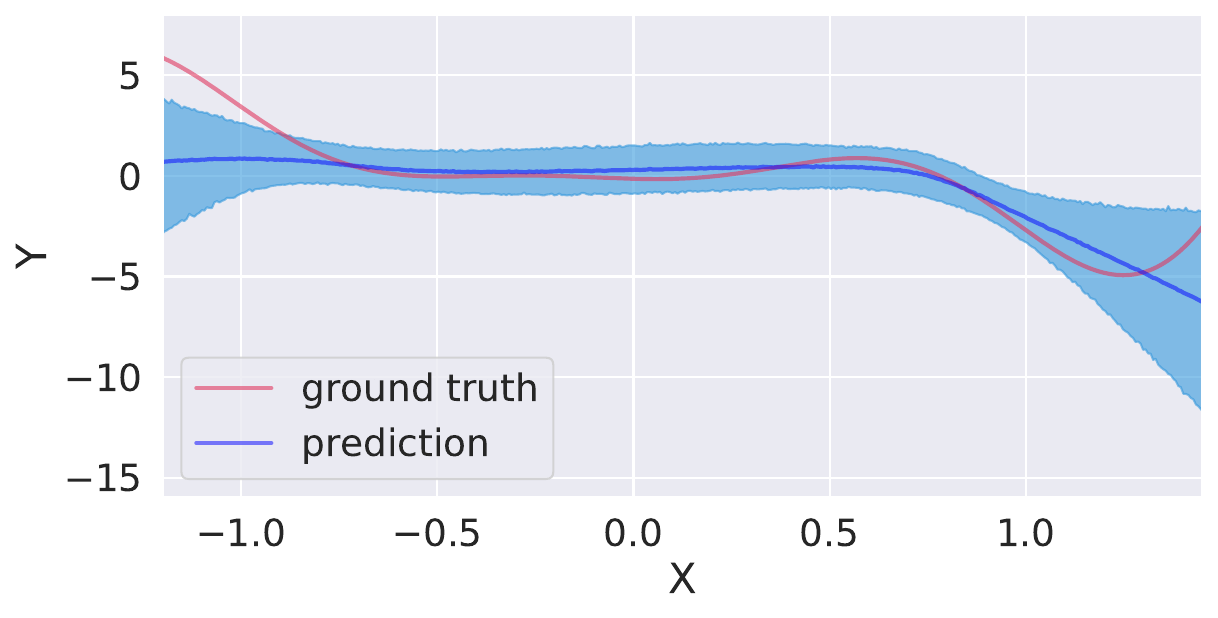}
    \caption{HMC with $2k$ samples}
    \label{fig:toy hmc}
    \end{subfigure}
    \caption{Uncertainty estimates for regression. The red line is the ground truth. The dark blue line shows the predictive mean. The shaded region is the $90\%$ confidence interval of the predictive distribution. 
    For the same number of samples, (b) \ciber is closer than (a) small-sample HMC to (c) a highly accurate but slow HMC with a large number of samples.
    See the Appendix for details. 
    }
    \label{fig:toy}
\end{figure*}

Marginalization is made possible by our observation that BMA reduces to weighted volume computation.
Certain classes of such problems can be solved exactly
by so-called weighted model integration~(WMI) solvers \citep{belle2015probabilistic}.
By closely approximating BMA with WMI, these solvers can provide accurate approximations to marginalization in BMA (cf.~Figure~\ref{fig:triangular surface}).
With this observation, we propose \ciber, a collapsed sampler that uses WMI for computing conditional distributions.
In the few-sample setting, \ciber delivers more accurate uncertainty estimates than the gold-standard Hamiltonian Monte Carlo~(HMC) method (cf.~Figure~\ref{fig:toy}). 
We further evaluate the effectiveness of \ciber on regression and classification benchmarks and show significant improvements over other Bayesian deep learning approaches in terms of both uncertainty estimation and accuracy.

\section{Bayesian Model Averaging as Weighted Volume Computation} 
\label{sec: preliminaries}

In \textbf{Bayesian Neural Networks (BNN)}, given a neural network $\nn_{\fw}$ parameterized by weights $\fw$, instead of doing inference with deterministic $\fw$ that
optimize objectives such as cross-entropy or mean squared error,
Bayesian learning infers a posterior $p(\fw \mid \data)$ over parameters $\fw$ after observing data $\data$.
During inference, this posterior distribution is then marginalized to produce final predictions.
This process is called \textbf{Bayesian Model Averaging (BMA)}. It can be seen as learning an ensemble of an infinite number of neural nets and aggregating their results.
Formally,
given input $\x$,
the posterior predictive distribution and the expected prediction for a regression task~are
\begin{equation}
    \label{eq:BMA}
    \begin{split}
    p(\y \mid \x) = \int p(\y \mid \x, \fw)~p(\fw \mid \data)\,d \fw, \qquad \text{and}
    \qquad
    \E_{p(\y \mid \x)}[\y] = \int \y~p(\y \mid \x) \,d \y.
    \end{split}
\end{equation}
For classification, the (most likely) prediction is the class $\argmax_{\y} p(\y \mid \x)$. 
BMA is intuitively attractive because it can be risky to base inference on a single neural network model. 
The marginalization in BMA gets around this issue by averaging over models according to a Bayesian posterior.

BMA requires approximations to compute posterior predictive distributions and expected predictions,
as the integrals in Equation~\ref{eq:BMA} are intractable in general. Deriving efficient and accurate approximations remains an active research topic~\citep{izmailov2021bayesian}.
We approach this problem by observing that
the marginalization in BMA with \relu neural networks can be cast as weighted volume computation~(WVC).
Later we show that it can be generalized to any neural network when combined with sampling.
In WVC, various tools exist for solving certain WVC problem classes~\citep{baldoni2014user,kolb2019pywmi,zeng2020probabilistic}.
This section reveals the connection between BMA and WVC. It opens up a new perspective for developing BMA approximations by leveraging WVC tools.

\begin{mydef}[WVC]
A weighted volume computation~(WVC) problem is defined by a pair $(\polytope, \wvcf)$ where a region $\polytope$ is a conjunction of arithmetic constraints
and weight $\wvcf: \polytope \rightarrow \R$ is an integrable function assigning weights to elements in $\polytope$. The task of WVC is to compute the integral $\int_{\polytope} \wvcf(\x)\,d\x$.
\end{mydef}

\subsection{A Warm-Up Example}
\label{sec:A Warm-Up Example}
Consider a simple yet relevant setting where
the predictive distribution $\predictivep$ is a Dirac delta distribution with zero mass everywhere except at $\nn_{\fw}(\x)$,
such that $\int \y~\predictivep \,d\y = \nn_{\fw}(\x)$.
\begin{exa}
\label{exa:avg relu}
Assume a model $\nn_{\fw}(x) = \relu(w \cdot x)$ 
with a uniform posterior over the parameter: $p(w~\mid~\data) = \frac{1}{6}$ with $w \in [-3, 3]$.
Let the input be $x = 1$.
For parameter $w \in [-3, 0]$, the model $\nn_{\fw}$ always predicts $0$, and otherwise (i.e., $w \in (0, 3]$), it predicts $w$.
Thus, the expected prediction (Equation~\ref{eq:BMA}) is
$
    \E_{p(\y \mid \x)}[\y]
    = \int_{\polytope_{\bot}} 0 \cdot \frac{1}{6} \,d\fw 
    + \int_{\polytope_{\top}} w \cdot \frac{1}{6} \,d\fw
$.
That is, a summation of two WVC problems $(\polytope_{\bot}, 0)$ and $(\polytope_{\top}, w / 6)$ with
$\polytope_{\bot} = (-3 \leq w \leq 0)$
and 
$\polytope_{\top} = (0 \leq w \leq 3)$. The BMA integral decomposes into WVC problems with different weights
due to the \relu activation.

\end{exa}

These WVC problems have easy closed-form solutions. 
This is no longer the case in the following.

\begin{exa}
\label{exa:gaussian avg relu}
Assume a model $\nn_{\fw}(x)$ and posterior distribution $p(w~\mid~\data)$ as in Example~\ref{exa:avg relu}.
Let the predictive distribution $p(y~\mid~x, w)$ be a Gaussian distribution 
$\gaussianp(y ; \nn_{\fw}(x), 1)$ with mean $\nn_{\fw}(x)$ and variance 1.
Given input $x = 1$, 
the expected prediction (Equation~\ref{eq:BMA}) is
\begin{equation*}
\begin{split}
    \E_{p(y \mid x = 1)}[y]
    = &\int_{\polytope_{\bot}} y \cdot \gaussianp(y \mid 0, 1) \cdot \frac{1}{6} \,d y\,d w
    + \int_{\polytope_{\top}} y \cdot \gaussianp(y \mid w, 1) \cdot \frac{1}{6} \,d y\,d w.
\end{split}
\end{equation*}
It is a summation of two WVC problems
with 
$\polytope_{\bot} = (-3 \leq w \leq 0) \land (y \in \R)$
and 
$\polytope_{\top} = (0 \leq w \leq 3) \land (y \in \R)$,
whose joint integral surface is shown in Figure~\ref{fig:gaussian surface}.
\end{exa}
These WVC problems do not admit closed-form solutions since they involve truncated 
Gaussian distributions.
Moreover, Figure~\ref{fig:gaussian surface} shows that computing BMA, even in such a low-dimensional parameter space, requires integration over non-convex and multi-modal functions.

\subsection{General Reduction of BMA to WVC}
Let model $\nn_{\fw}$ be a \relu neural net. 
Denote the set of inputs to its \relu activations by 
$\allrelu = \{ \reluinput_i \}_{i=1}^R$, where each $\reluinput_i$ is a linear combination of weights.
For a given input $\x$, the parameter space is partitioned by whether each \relu activation outputs zero or not. This gives the WVC reduction 
\begin{align*}
p(\y \mid \x) &= \sum_{\bidx \in \{0, 1\}^R } \int_{\polytope_{\bidx}} p(\y \mid \x, \fw)~p(\fw \mid \data)\,d \fw,
\end{align*}
where $\bidx$ is a binary vector. The region $\polytope_{\bidx}$ is defined as $\land_{i=1}^R \ell_i$ where arithmetic constraint $\ell_i$ is 
$\reluinput_i \geq 0$ if $\bidx_i = 1$ and $\reluinput_i \leq 0$ otherwise.
The expected prediction $\E_{p(\y \mid \x)}[\y]$ is analogous but includes an additional factor and variable of integration $y$ in each WVC problem.

This general reduction, however, is undesirable since it amounts to a brute-force enumeration that implies a complexity exponential in the number of \relu activations.
Moreover, not all WVC problems resulting from this reduction are amenable to existing solvers.
We will therefore appeal to a framework called weighted model integration~(WMI) that allows for a compact representation of these WVC problems, and a characterization of their tractability for WMI solvers~\citep{kolb2019pywmi}.
This inspires us to approximate BMA by first reducing it to WVC problems and further 
closely approximating those with tractable WMI problems.

\section{Approximating BMA by WMI}
\label{sec: exact BMA over relu}

WMI is a 
modeling and inference framework that supports integration
in the presence of logical and arithmetic constraints~\citep{belle2015probabilistic,BelleUAI15}. 
Various WMI solvers have been proposed in recent years~\citep{kolb2019pywmi}, ranging from general-purpose ones to others that assume some problem structures to gain scalability. 
However, even with the reduction from BMA to WVC from the previous section, WMI solvers are not directly applicable.
Existing solvers have two main limitations:
(i)~feasible regions need to be defined by Boolean combinations of linear arithmetic constraints, and 
(ii)~weight functions need to be polynomials.
In this section, we show that these issues can be bypassed using a motivating example of how to form a close approximation to BMA using WMI.

In WMI, the feasible region is defined by \emph{satisfiability modulo theories} (SMT) constraints~\citep{barrett2010smt}:
an SMT formula is a (typically quantifier-free) expression containing both propositional and theory literals connected with logical connectives;
the theory literals
are often restricted to \emph{linear real arithmetic}, where literals are of the form $(\mathbf{c}^T \cvars \le b)$ with variable $\cvars$ and constants $\mathbf{c}^T$ and $b$.
\begin{exa}
\label{exa: relu smt formula}
The \relu model 
$\nn_{\fw}(x)$
of Example~\ref{exa:avg relu} can be encoded as an SMT formula (see box).
\vspace{-0.5em}
{\makeatletter
\let\par\@@par
\par\parshape0
\everypar{}
\begin{wrapfigure}{r}{.45\textwidth}
\vspace{-2em}
\begin{equation*}
\boxed{
    \theory_{\relu} = \left\{ 
    \begin{array}{l}
        W \cdot x > 0 \Rightarrow Z = W \cdot x \\
        W \cdot x \leq 0 \Rightarrow Z = 0
    \end{array}
    \right.
}
\end{equation*}
\end{wrapfigure}
The curly bracket denotes logical conjunction, the symbol $\Rightarrow$ is a logical implication, variable $W$ is the weight, and variable $Z$ denotes the model output.
\par}%
\end{exa}
The encoding of \relu neural networks into SMT formulas is explored in existing work to enable verification of the behavior of neural networks and provide formal guarantees~\citep{katz2017reluplex,huang2017safety,SivaramanNeurIPS20}. 
We propose to use
this encoding to define the feasible region of WMI problems.
Let $\x \models \theory$ denote the satisfaction of an SMT formula $\theory$ by an assignment $\x$, and $\id{\x \models \theory}$ be its corresponding indicator function. We formally introduce WMI next.

\begin{mydef}{(WMI)}
Let $\X$ be a set of continuous random variables.
A \textit{weighted model integration} problem is a pair $\probsmt = (\theory, \Weights)$,
where
$\theory$
is an SMT formula over $\X$
and $\Weights$ is a set of per-literal weights defined as $\Weights = \{\weight_{\literal}\}_{\literal \in \literals}$,
where $\literals$ is a set of \smt literals and each $\weight_{\literal}$ is a function defined over variables in literal $\literal$.
The task of weighted model integration is to compute
\begin{equation*}
    \WMI(\theory, \Weights) = \int_{\x \models \theory} 
    \prod_{\literal \in \literals} \weight_{\literal}(\x)^{\id{\x \models \literal}}
    \, d\x.
\end{equation*}
\end{mydef}
That is, the task is to integrate over the weighted assignments of $\X$ that satisfy the \smt formula $\theory$.\footnote{In the literature, WMI is defined on mixed discrete-continuous domains. Since we only work with WMI problems over continuous variables, we ignore the discrete ones in the definition for succinctness.}

An approximation to the BMA of Example~\ref{exa:gaussian avg relu} can be achieved with WMI using the following four~steps:

\textbf{Step 1. Encoding model $\nn_{\fw}(x)$.}
This has been shown as the SMT formula $\theory_{\relu}$ in 
Example~\ref{exa: relu smt formula}.

\textbf{Step 2. Encoding posterior distribution $p(w \mid \data)$.}
The uniform distribution $p(w~\mid~\data) = \frac{1}{6}$ with $w \in [-3, 3]$ can be encoded as a WMI problem pair
$(\theory_{\posterior}, \Weights_{\posterior})$ as follows:
\begin{shrinkeq}{0ex}
\begin{align*}
\boxed{
~~\theory_{\posterior} = -3 \leq W \leq 3 
\qquad
\Weights_{\posterior} =
    \left\{
    \weight_{\ell}(W) = \frac{1}{6}
    ~~\textit{with}~~
    \ell = \true
    \right\}
}~~
\end{align*}
\end{shrinkeq}

\textbf{Step 3. Approximate encoding of predictive distribution $p(y \mid w, x)$.}
Recall that $p(y \mid w, x) = \gaussianp(y; \nn_{\fw}(x), 1)$ is Gaussian, which cannot be handled by existing WMI solvers. 
To approximate it with polynomial densities,
we simply use a triangular distribution encoded as a WMI problem pair:
\begin{align*}
\boxed{
~~\theory_{\prediction} = 
\left\{ 
\begin{array}{l}
    Y \leq Z + \trir \\
    Y \geq Z - \trir
\end{array}
\right.
\qquad
\Weights_{\prediction} =
\left\{
\begin{array}{l}
    \weight_{\literal_1}(Y, Z) = \frac{1 - Y + Z}{\trir}
    ~~\textit{with}~~ \literal_1 = Y \geq Z \\
    \weight_{\literal_2}(Y, Z) = \frac{1 + Y - Z}{\trir}
    ~~\textit{with}~~ \literal_2 = Y < Z
\end{array}
\right\}
~~}
\end{align*}

\begin{wrapfigure}[5]{r}{0.30\textwidth}
\centering
\vspace{-0.0em}
    \includegraphics[trim=0cm 1cm 0cm 0cm, width=0.30\textwidth, clip]{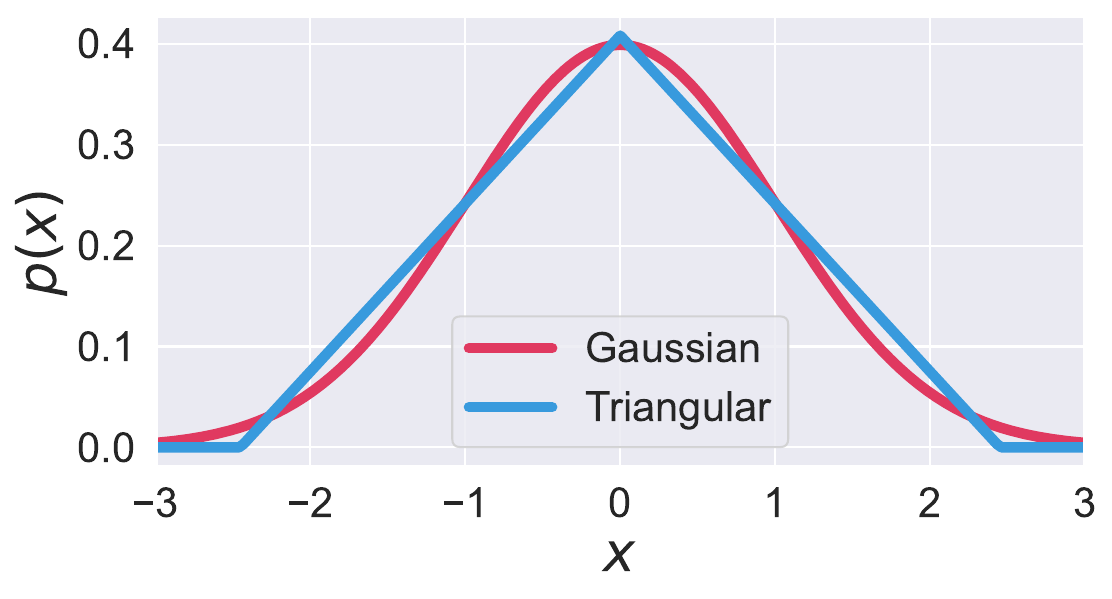}
\end{wrapfigure}
In this encoding, $\trir$ is a constant that defines the shape of the triangular distribution. It is obtained by minimizing the $L2$ distance between a standard normal distribution and the symmetric triangular distribution.
We visualize this approximation in the right figure. 

\textbf{Step 4. Approximating BMA by calling WMI solvers.}
With the above encodings,
the predictive posterior $p(y \mid x)$ (Equation~\ref{eq:BMA}) can be computed using two calls to a WMI solver. For example, the uncertainty of a prediction $y = 1$ for input $x = 1$ is
\begin{align*}
    p(y = 1 \mid x = 1) 
    ~=~ 
        \WMI(\theory \land (Y = 1), \Weights)
        ~/~
        \WMI(\theory, \Weights)
    ~=~ 
    0.164 ~/~ 1,
\end{align*}
where $\theory = \theory_{\relu} \land \theory_{\posterior} \land \theory_{\prediction}$ and $\Weights = \Weights_{\posterior} \cup \Weights_{\prediction}$.
Similarly, the expected prediction $\E_{p(y \mid x = 1)}[y]$ (Equation~\ref{eq:BMA}) can be computed using two calls to a WMI solver:
\begin{align*}
    \E_{p(y \mid x = 1)}[y]
    ~=~
        \WMI(\theory, \Weights^*)
    ~/~
        \WMI(\theory, \Weights)
    ~=~ 0.752 ~/~ 1,
\end{align*}
where $\Weights^* = \Weights \cup \{\weight_{\literal}(Y) = Y ~~\textit{with}~~ \literal = \true \}$.
The above formulations also work for unnormalized distributions since the WMI in the denominator serves to compute the partition function.

We visualize the integral surface of the resulting approximate BMA problem in Figure~\ref{fig:triangular surface}. It is very close to the integral surface of the original BMA problem in Figure~\ref{fig:gaussian surface}.
However, it can be exactly integrated using existing WMI solvers while the original one does not admit such solutions.
Next, we show how this process can be generalized to a scalable and accurate approximation of BMA.

\section{\ciber: Collapsed Inference for Bayesian Deep Learning via WMI}
\label{sec: ciber}

Given a BNN with a large number of weights,
naively approximating it by WMI problems can lead to computational issues,
since it involves doing integration over polytopes in arbitrarily high dimensions and this is known to be \#P-hard~\citep{valiant1979complexity,de2012software,zeng2020probabilistic}. 
Further, weights involved with non-\relu activation might not be amenable to the WMI encoding.
To tackle these issues, we propose to use collapsed samples to combine the strengths from two worlds: the scalability and flexibility from sampling and the accuracy from WMI solvers.

\begin{mydef}{(Collapsed BMA)}
Let $(\W_{\sampling}, \W_{\collapse})$ be a partition of parameters $\W$.
A collapsed sample is a tuple $(\fw_{\sampling}, q)$,
where $\fw_{\sampling}$ is an assignment to the sampled parameters $\W_{\sampling}$ and $q$ is a representation of the conditional posterior $p(\W_{\collapse} \mid \fw_{\sampling}, \data)$ over the collapsed parameter set $\W_{\collapse}$.
Given collapsed samples $\mathcal{S}$,
\emph{collapsed BMA} estimates the predictive posterior and expected prediction~as
\begin{equation}
\label{eq: collapsed BMA}
\begin{split}
    p(\y \mid \x) 
    &\approx 
    \frac{1}{|\mathcal{S}|}
    \sum_{(\fw_{\sampling}, q) \in \mathcal{S}} 
    \left[\int p(\y \mid \x, \fw)~q(\fw_{\collapse}) \,d\fw_{\collapse}\right], \text{ and}\\
    \E_{p(\y \mid \x)}[\y] 
    &\approx
    \frac{1}{|\mathcal{S}|}
    \sum_{(\fw_{\sampling}, q) \in \mathcal{S}} 
    \left[\int \y~p(\y \mid \x, \fw)~q(\fw_{\collapse}) \,d\fw_{\collapse}~d\y \right].
\end{split}
\end{equation}
\end{mydef}
The size of the collapsed set $\W_{\collapse}$ determines the trade-off between scalability and accuracy. The more parameters in the collapsed set, the more accurate the approximation to BMA is. The fewer parameters in $\W_{\collapse}$, the more efficient the computations of the integrals are since the integration is performed in a lower-dimensional space.
Later in our experiments,
we choose a subset of weights at the last or second-to-last hidden layer of the neural networks to be the collapsed set. This choice is known to be effective in capturing uncertainty as shown in \citet{kristiadi2020being, snoek2015scalable}.

To develop an algorithm to compute collapsed BMA, we are faced with two main design choice questions:
\textbf{(Q1)} how to sample $\fw_{\sampling}$ from the posterior? %
\textbf{(Q2)} what should be the representation of the conditional posterior $q$ such that the integrals in Equation~\ref{eq: collapsed BMA} can be computed exactly?
Next, we provide our answers to these two questions that together give our proposed algorithm \ciber.

\subsection{Approximation to Posteriors}
For \textbf{(Q1)}, we follow \citet{maddox2019simple} and sample from the stochastic gradient descent~(SGD) trajectory after convergence and use the information contained in SGD trajectories to efficiently approximate the posterior distribution over the parameters of the neural network, leveraging the interpretation of SGD as approximate Bayesian inference~\citep{mandt2017stochastic,chen2020statistical}.
Given a set of parameter samples $\sampleW$ from the SGD trajectory,
the sample set is defined as $\sampleW_{\sampling} = \{\fw_{\sampling} \mid \fw \in \sampleW\}$.
For each assignment $\fw_{\sampling}$, an approximation $q(\W_{\collapse})$ to the conditional posterior $p(\W_{\collapse} \mid \fw_{\sampling}, \data)$ is necessary since the posteriors induced by SGD trajectories are implicit.
Next, we discuss the choice of approximation 
to the conditional posterior that is amenable to WMI.

\subsection{Encoding into WMI Problems}
\label{sec: probs smt encoding}

As shown in Section~\ref{sec: exact BMA over relu}, if a BNN can be encoded as a WMI problem,
the posterior predictive distribution and the expected prediction, which involve marginalization over the parameter space, can be computed exactly using WMI solvers.
This inspires us to use the WMI framework as the closed-form representation for the conditional posteriors of parameters.
The main challenge is how to approximate the integrand in Equation~\ref{eq: collapsed BMA} using an SMT formula and a polynomial weight function in order to obtain a WMI problem amenable to existing solvers.

\textit{For the conditional posterior approximation $q(\W_{\collapse})$}, 
we choose it to be a uniform distribution that can be encoded into a WMI problem as $\probsmt_{\posterior} = (\theory_{\posterior}, \Weights_{\posterior})$ with the SMT formula being $\theory_{\posterior} = \land_{i \in \collapse} ~(l_i \leq W_i \leq u_i)$ and
weights being $\Weights_{\posterior} = \left\{ \weight_{\ell}(\W_{\collapse}) = 1 \mid \ell = \true \right\}$, where $l_i$ and $u_i$ are domain lower and upper bounds for the uniform distribution respectively.
While seemingly over-simplistic, this choice of approximation to the conditional posterior is sufficient to robustly deliver surprisingly strong empirical performance as shown in Section~\ref{sec: experiment}.
The intuition is that uniform distributions are better than a few samples.
We further illustrate this point by comparing the predictive distributions of CIBER and HMC in a few-sample setting. Figure~\ref{fig:toy} shows that even with the same $10$ samples drawn from the posterior distribution, since CIBER further approximates the $10$ samples with a uniform distribution, it yields a predictive distribution closer to the ground truth than HMC, indicating that using a uniform distribution instead of a few samples forms a better approximation.

\textit{For the choice of predictive distribution $p(\y \mid \x, \fw)$}, 
we propose to use piecewise polynomial densities.
Common predictive distributions 
can be approximated by polynomials up to arbitrary precision in theory by the Stone–Weierstrass theorem~\citep{de1959stone}. For regression,
the de facto choice is Gaussian and
we propose to use triangular distribution as the approximation,
i.e.,
$p(\y \mid \x, \fw) 
= \frac{1}{\radius} - \frac{1}{\radius^2}|\y - \nn_{\fw}(\x)|$, with domain $|\y - \nn_{\fw}(\x)| \leq \radius$, 
and $\radius := \trir \sqrt{\variance(\x)}$
where the constant $\trir$ parameterizes the triangular distribution as described in Section~\ref{sec: exact BMA over relu}.
Here, $\variance(\x)$ is the variance estimate, which can be a function of input $\x$ depending on whether the BNN is homoscedastic or heteroscedastic.
Then $p(\y \mid \x, \fw)$ can be encoded into WMI as:
\begin{align*}
\boxed{
\resizebox{.99\textwidth}{!}{$
\begin{array}{l}
\theory_{\prediction} = 
\left\{ 
\begin{array}{l}
    Y - \nn_{\fw}(\x) \leq \radius \\ 
    Y - \nn_{\fw}(\x) \geq -\radius
\end{array}
\right.
~~
\Weights_{\prediction} =
\left\{
\begin{array}{l}
    \weight_{\literal_1}(Y, \W_{\collapse}) = \frac{1}{\radius} - \frac{Y - \nn_{\fw}(\x)}{\radius^2}
    ~~\textit{with}~~ \ell_1 = (Y > \nn_{\fw}(\x))\\
    \weight_{\literal_2}(Y, \W_{\collapse}) = \frac{1}{\radius} - \frac{\nn_{\fw}(\x) - Y}{\radius^2}
    ~~\textit{with}~~ \ell_2 = (\nn_{\fw}(\x) > Y)\\
\end{array}
\right\}
\end{array}
$}
}
\end{align*}
Similar piecewise polynomial approximations are adopted for classification tasks when the predictive distributions are induced by softmax functions. Those details are presented in the Appendix.

\subsection{Exact Integration in Collapsed BMA}
\label{sec: Exact Integration in Collapsed BMA}
By encoding the collapsed BMA into WMI problems,
we are ready to answer \textbf{(Q2)}, i.e., how to perform exact computation of the integrals 
shown in Equation~\ref{eq: collapsed BMA}.

\begin{pro}
\label{prop: collapsed}
Let the SMT formula $\theory = \theory_{\relu} \land \theory_{\posterior} \land \theory_{\prediction}$, and the set of weights $\Weights = \Weights_{\posterior} \cup \Weights_{\prediction}$ as defined in Section~\ref{sec: probs smt encoding}. Let the set of weights
$\Weights^* = \Weights \cup \{\weight_{\literal}(Y) = Y ~~\textit{with}~~ \literal = \true \}$.
The integrals in collapsed BMA (Equation~\ref{eq: collapsed BMA}) can be computed by WMI solvers as
\begin{align*}
    \int p(\y \mid \x, \fw)~q(\fw_{\collapse}) \,d\fw_{\collapse} 
    &=
    \WMI(\theory \land (\bm{Y} = \y), \Weights)
    ~/~
    \WMI(\theory, \Weights)
, \text{ and}\\
    \int \y~p(\y \mid \x, \fw)~~q(\fw_{\collapse}) \,d\fw_{\collapse}~d\y
    &=
        \WMI(\theory, \Weights^*)
    ~/~
        \WMI(\theory, \Weights).
\end{align*}
\end{pro}
With both questions \textbf{(Q1)} and \textbf{(Q2)} answered, we summarize our proposed algorithm \ciber in Algorithm~\ref{algo: ciber} in the Appendix. 
To quantitatively analyze how close the approximation delivered by \ciber is to the ground-truth BMA, we consider the following experiments with closed-form BMA.

\begin{minipage}[t]{.53\textwidth}
    \centering
    \includegraphics[height=0.12\textheight]{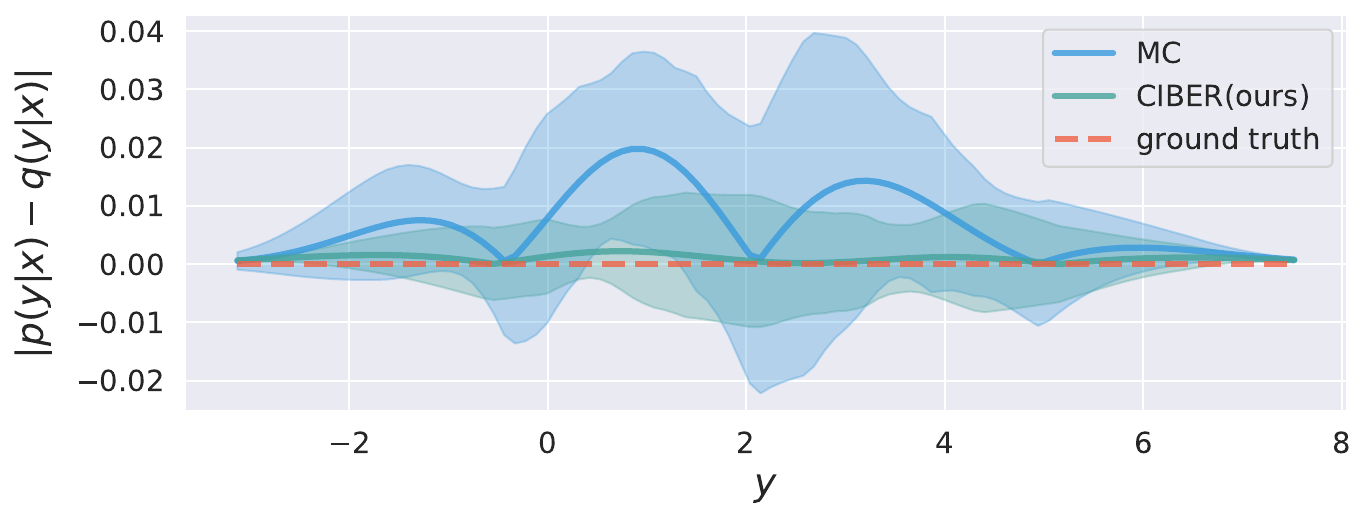}
    \captionof{figure}{Posterior predictive distributions in Bayesian linear regression.
    The $y$-axis shows the absolute difference between an estimated predictive distribution $p(y \mid \mathbf{x})$ and the ground-truth predictive distribution $q(y \mid \mathbf{x})$.
    Shaded regions are the $95\%$ confidence interval.
    Best viewed in color.
    }
    \label{fig:true bma 10}
\end{minipage}
\hfill
\begin{minipage}[t]{.45\textwidth}
    \centering
    \includegraphics[height=0.12\textheight]{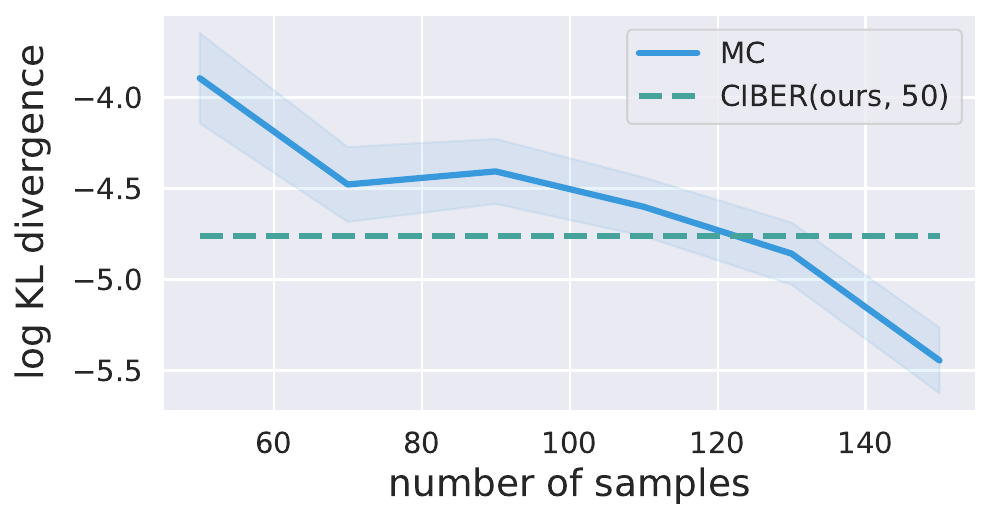}
    \captionof{figure}{KL divergence 
    in Bayesian linear regression.
    The $x$-axis shows the number of samples the MC method uses for estimations, ranging from $50$ to $150$. 
    The blue curve shows the MC method, and the green dashed curve shows CIBER using $50$ samples. 
    }
    \label{fig:ablation}
\end{minipage}

\textbf{Regression.}
We consider a Bayesian linear regression setting where exact sampling from the posterior distribution is available. Both the likelihood and the weight posterior are Gaussian such that the ground-truth posterior predictive distribution is Gaussian as well.
With samples drawn from the weight posterior, \ciber approximates the samples with a uniform distribution as posterior $p(\fw|\data)$ and further approximates the likelihood with a triangular distribution such that the integral $p(\y|\x, \data) = \int p(\y|\x, \fw) p(\fw|\data) \,d\fw$ can be computed exactly by WMI.

We first evaluate the posterior predictive distribution estimated by \ciber and Monte Carlo~(MC) method, using the same five samples drawn from the weight posterior.
Results averaged over $10$ trials are shown in Figure~\ref{fig:true bma 10} where the estimations by CIBER are much closer to the ground truth posterior predictive distribution than those by the MC method.
Further, the averaged KL divergence between the ground truth and CIBER is $0.069$ while the one for MC estimations is $0.130$, again indicating that CIBER yields a better BMA approximation in the few-sample setting.

We further explore the question of how many samples the MC method needs to match the performance of CIBER. 
The performances of both approaches are evaluated using KL divergence between the ground-truth posterior distribution and the estimated one, averaged over 10 trials. The result is shown in Figure~\ref{fig:ablation} where the dashed green line shows the performance of CIBER with $50$ samples and the blue curve shows the performance of MC with an increasing number of samples. As expected, the MC method yields lower KL divergence as the number of samples increases; however, it takes more than $100$ samples to match CIBER, indicating its low sample efficiency and that developing efficient and effective inference algorithms such as CIBER for estimating BMA is a meaningful question.

\textbf{Classification.}
For analyzing classification performance, \citet{kristiadiposterior} propose to compute the integral $I = \int \sigma(\nn_*) \gaussianp(\nn_*) \, d \nn_*$ with $\sigma$ being the sigmoid function and $\nn_* = \nn(\x^*; \fw)$ that amounts to the posterior predictive distribution. We consider a simple case with $\nn(\x; \fw) = \fw\cdot\x$ such that the ground-truth integral can be obtained. With a randomly chosen $\x$, the ground-truth integral is $I = 0.823$. 
The integral estimated by \ciber is $I_{C} = 0.826$ while the MC estimate is $I_{\mathit{MC}} = 0.732$. That is, \ciber gives an estimate with a much lower error than the MC estimation error, indicating that \ciber is able to deliver high-quality approximations in classification tasks.

\section{Related Work}
\label{sec: related work}
\textbf{Bayesian Deep Learning.\ } Bayesian inference over deep neural networks~\citep{mackay1992evidence} is proposed to fix the issue that 
deep learning models give poor uncertainty estimations and suffer from overconfidence~\citep{nguyen2015deep,hein2019relu,meronen2023fixing,meronen2021periodic}.
Some methods use samples from SGD trajectories to approximate the implicit true posteriors similar to us:
\citet{izmailov2020subspace} (SI) proposes to perform Bayesian inference in a subspace of the parameter space spanned by a few vectors derived from principal component analysis 
(PCA+ESS(SI)) or variational inference (PCA+VI(SI)); SWAG~\citep{maddox2019simple} proposes to approximate the full parameter space using an approximate Gaussian posterior whose mean and covariance are from a partial SGD trajectory with a modified learning rate scheduler.

Some other approaches using approximate posteriors include
MC Dropout (MCD)~\citep{gal2015bayesian,gal2016dropout} which is one of the Bayesian dropout methods and recently, one of its modifications called Variational Structured Dropout~(VSD)~\citep{nguyen2021structured} using variational inference is proposed.
Other state-of-the-art approximate BNN inference methods including deterministic variational inference (DVI)~\citep{wu2018deterministic},
deep Gaussian processes (DGP)~\citep{bui2016deep} with Gaussian process layers 
and variational inference (VI)~\citep{kingma2013auto}.
Closely related to DGP is the deep kernel process~\citep{aitchison2021deep} that writes DGPs as deep Wishart processes.

\textbf{WMI Solvers.\ }
WMI generalizes weighted model counting (WMC)~\cite{sang2005performing},
a state-of-the-art inference approach in many discrete probabilistic models,
from discrete to mixed discrete-continuous domains~\cite{belle2015probabilistic,BelleUAI15}. 
Recent research on WMI includes its tractability~\citep{zeng2020probabilistic,zengparameter,abboud2020approximability}
and the advancements in WMI solver development.
Existing exact WMI solvers for arbitrarily structured problems include DPLL-based search with numerical~\cite{belle2015probabilistic,morettin2017efficient,morettin2019advanced} or symbolic integration~\cite{braz2016probabilistic} and compilation-based algorithms~\cite{kolb2018efficient,zuidberg2019exact,derkinderen2020ordering} that use extended algebraic decision diagrams~(XADDs)~\citep{sanner2012symbolic2} as a compilation target which is a powerful tool for inference on mixed domains~\citep{sanner2012symbolic,zamani2012symbolic}.
Some exact WMI solvers aiming to improve efficiency for a certain class of models are proposed such as SMI~\citep{zeng2019efficient} and MP-WMI~\citep{zeng2020scaling} which are greatly scalable for WMI problems that satisfy certain structural constraints.
Approximate solvers are also proposed including sampling-based ones~\citep{zuidberg2020monte} and relaxation-based ones~\citep{zeng2020probabilistic,ZengDECODEML20}.
Recent WMI efforts converge in the \texttt{pywmi} library~\citep{kolb2019pywmi}.
The SMT formulas considered in this work can be seen as distributional constraints on continuous domains. 
There is also plenty of work in neuro-symbolic AI exploring the integration of discrete constraints
into neural networks models including the architectures~\citep{AhmedNeurIPS22, SIMPLE} and the loss~\citep{xu2018semantic, Ahmed2022neuro, AhmedAAAI22}.

\begin{table*}
\caption{
Average test log likelihood for the small UCI regression task.
}
\centering
\begin{adjustbox}{max width=0.99\textwidth}
\sc
{\setlength\doublerulesep{1pt} 
\begin{tabular}{@{}llllll@{}}
\toprule[1pt]\midrule[0.3pt]
& Boston & Concrete  & Yacht   & Naval  & Energy \\
\midrule
\ciber(second) & \textbf{-2.471 $\pm$ 0.140} & -2.975 $\pm$ 0.102 & -0.678 $\pm$ 0.301 & 7.276 $\pm$ 0.532 & \underline{\textbf{-0.716 $\pm$ 0.211}} \\
\ciber(last) & \textbf{-2.471 $\pm$ 0.140} & \underline{\textbf{-2.959 $\pm$ 0.109}} & -0.687 $\pm$ 0.301 & \underline{\textbf{7.482 $\pm$ 0.188}} & \underline{\textbf{-0.716 $\pm$ 0.211}} \\
SWAG & -2.761 $\pm$ 0.132 & -3.013 $\pm$ 0.086 & -0.404 $\pm$ 0.418 & 6.708 $\pm$ 0.105 & -1.679 $\pm$ 1.488 \\
PCA+ESS (SI) & -2.719 $\pm$ 0.132 & -3.007 $\pm$ 0.086 & \underline{\textbf{-0.225 $\pm$ 0.400}} & 6.541 $\pm$ 0.095 & -1.563 $\pm$ 1.243 \\
PCA+VI (SI) & -2.716 $\pm$ 0.133 & -2.994 $\pm$ 0.095 & -0.396 $\pm$ 0.419 & 6.708 $\pm$ 0.105 & -1.715 $\pm$ 1.588 \\
\hline
SGD & -2.752 $\pm$ 0.132 & -3.178 $\pm$ 0.198 & -0.418 $\pm$ 0.426 & 6.567 $\pm$ 0.185 & -1.736 $\pm$ 1.613 \\
DVI & -2.410 $\pm$ 0.020 & -3.060 $\pm$ 0.010 & -0.470 $\pm$ 0.030 & 6.290 $\pm$ 0.040 & -1.010 $\pm$ 0.060 \\
DGP & \underline{-2.330 $\pm$ 0.060} & -3.130 $\pm$ 0.030 & -1.390 $\pm$ 0.140 & 3.600 $\pm$ 0.330 & -1.320 $\pm$ 0.030 \\
VI & -2.430 $\pm$ 0.030 & -3.040 $\pm$ 0.020 & -1.680 $\pm$ 0.040 & 5.870 $\pm$ 0.290 & -2.380 $\pm$ 0.020 \\
MCD & -2.400 $\pm$ 0.040 & -2.970 $\pm$ 0.020 & -1.380 $\pm$ 0.010 & 4.760 $\pm$ 0.010 & -1.720 $\pm$ 0.010 \\
VSD & -2.350 $\pm$ 0.050 & -2.970 $\pm$ 0.020 & -1.140 $\pm$ 0.020 & 4.830 $\pm$ 0.010 & -1.060 $\pm$ 0.010 \\
\midrule[0.3pt]\bottomrule[1pt]
\end{tabular}
}
\end{adjustbox}
\label{tab: small likelihood}
\end{table*}
\section{Experiments}
\label{sec: experiment}
We conduct experimental evaluations of our proposed approach \ciber~\footnote[1]{Code and experiments are available at \texttt{https://github.com/UCLA-StarAI/CIBER}.} on regression and classification benchmarks and compare its performance on uncertainty estimation as well as prediction accuracy with a wide range of baseline methods. More experimental details are presented in the Appendix.
\begin{table*}[t]
\centering
\caption{
Average test log likelihood for the large UCI regression task.
}
\begin{adjustbox}{max width=0.99\textwidth}
\sc
{\setlength\doublerulesep{1pt} 
\begin{tabular}{@{}lllllll@{}}
\toprule[1pt]\midrule[0.3pt]
& Elevators & KeggD  & KeggU   & Protein  & Skillcraft & Pol \\
\midrule
\ciber(second) & -0.378 $\pm$ 0.026 & \underline{\textbf{1.245 $\pm$ 0.090}} & \underline{\textbf{1.125 $\pm$ 0.269}} & -0.720 $\pm$ 0.036 & -1.003 $\pm$ 0.035 & \underline{\textbf{2.555 $\pm$ 0.115}}\\
\ciber(last) & -0.371 $\pm$ 0.023 & 1.178 $\pm$ 0.088  & 0.964 $\pm$ 0.231 & -0.720 $\pm$ 0.036 & \underline{\textbf{-1.001 $\pm$ 0.032}} & 2.506 $\pm$ 0.150 \\
SWAG & -0.374 $\pm$ 0.021 & 1.080 $\pm$ 0.035 & 0.749 $\pm$ 0.029 & \underline{\textbf{-0.700 $\pm$ 0.051}} & -1.180 $\pm$ 0.033 & 1.533 $\pm$ 1.084 \\
PCA+ESS (SI) & -0.351 $\pm$ 0.030 & 1.074 $\pm$ 0.034 & 0.752 $\pm$ 0.025 & -0.734 $\pm$ 0.063 & -1.181 $\pm$ 0.033 & -0.185 $\pm$ 2.779 \\
PCA+VI (SI) & \underline{\textbf{-0.325 $\pm$ 0.019}} & 1.085 $\pm$ 0.031 & 0.757 $\pm$ 0.028 & -0.712 $\pm$ 0.057 & -1.179 $\pm$ 0.033 & 1.764 $\pm$ 0.271 \\
\hline
SGD & -0.538 $\pm$ 0.108 & 1.012 $\pm$ 0.154 & 0.602 $\pm$ 0.224 & -0.854 $\pm$ 0.085 & -1.162 $\pm$ 0.032 & 1.073 $\pm$ 0.858 \\
OrthVGP & -0.448 & 1.022 & 0.701 & -0.914 & --- & 0.159 \\
NL & -0.698 $\pm$ 0.039 & 0.935 $\pm$ 0.265 & 0.670 $\pm$ 0.038 & -0.884 $\pm$ 0.025 & -1.002 $\pm$ 0.050 & -2.840 $\pm$ 0.226 \\
\midrule[0.3pt]\bottomrule[1pt]
\end{tabular}
}
\end{adjustbox}
\label{tab: large likelihood}
\end{table*}

\subsection{Regression on Small and Large UCI Datasets}
\label{sec: Regression on Small and Large UCI Datasets}

We experiment on $5$ small UCI datasets: 
\emph{boston}, \emph{concrete}, \emph{yacht}, \emph{naval} and \emph{energy}.
We follow the setup of \citet{izmailov2020subspace} and use a fully connected network with a single hidden layer and $50$ units with \relu activations. 
We further experiment on $6$ large UCI datasets:
\emph{elevators}, \emph{keggdirected}, \emph{keggundirected}, \emph{pol}, \emph{protein} and \emph{skillcraft}.
We use a feedforward network with five hidden layers of sizes $[1000, 1000, 500, 50, 2]$ and \relu activations on all datasets except \emph{skillcraft}. For \emph{skillcraft}, a smaller architecture is adopted with four hidden layers of size $[1000, 500, 50, 2]$.
All models have two outputs for the prediction and the heteroscedastic variance respectively.

We run \ciber with two different ways of choosing the collapsed parameter set:
\emph{\ciber(last)} chooses all the weights at the last layer to be the collapsed set;
\emph{\ciber(second)} chooses three out of all the weights at the second-to-last layer to be the collapsed set.
The heuristic we use for choosing the weights is to look into the sampled weights from SGD trajectories to see which ones have the greatest variance.
The intuition is that a greater variance indicates that the weight is
prone to have greater uncertainty and thus one might want to perform a more accurate inference over it.

\textbf{Baselines.}
We compare \ciber to the state-of-the-art approximate BNN inference methods. We separate these methods into two categories:
those sampling from SGD trajectories as approximate posteriors, which includes
SWAG~\citep{maddox2019simple}, PCA+ESS (SI) and PCA+VI (SI)~\citep{izmailov2020subspace}, vs.\ those who do not, which includes the SGD baseline,  deterministic variational inference (DVI)~\citep{wu2018deterministic}, Deep Gaussian Processes (DGP)~\citep{bui2016deep}, variational inference (VI)~\citep{kingma2013auto}, MC Dropout~(MCD)~\citep{gal2015bayesian,gal2016dropout},
and variational structured dropout~(VSD)~\citep{nguyen2021structured}. These methods achieved state-of-the-art performance on the small UCI datasets. We also compare to baselines Bayesian final layers~(NL)~\citep{riquelme2018deep}, deep kernel learning~(DKL)~\citep{wilson2016deep}, orthogonally decoupled variational GPs (OrthVGP)~\citep{salimbeni2018orthogonally} and Fastfood approximate kernels~(FF) \citep{yang2015carte}, which have achieved state-of-the-art performance on the large UCI datasets.

\begin{table*}[t]
\centering
\caption{Average test performance for image classification tasks on CIFAR-10 and CIFAR-100. 
}
\begin{adjustbox}{max width=0.99\textwidth}
\sc
{\setlength\doublerulesep{1pt} 
\begin{tabular}{@{}lllllll@{}}
\toprule[1pt]\midrule[0.3pt]
Metric & \multicolumn{2}{c}{NLL} & \multicolumn{2}{c}{ACC} & \multicolumn{2}{c}{ECE}\\
\midrule
Dataset & CIFAR-10 & CIFAR-100 & CIFAR-10 & CIFAR-100 & CIFAR-10 & CIFAR-100 \\
\midrule
CIBER & \textbf{0.1927 $\pm$ 0.0029}  & \textbf{0.9193 $\pm$ 0.0027}  & \textbf{93.64 $\pm$ 0.09}  & \textbf{74.71 $\pm$ 0.18}  & 0.0130 $\pm$ 0.0011  & \textbf{0.0168 $\pm$ 0.0025}  \\
SWAG & 0.2503 $\pm$ 0.0081  & 1.2785 $\pm$ 0.0031  & 93.59 $\pm$ 0.14  & 73.85 $\pm$ 0.25  & 0.0391 $\pm$ 0.0020  & 0.1535 $\pm$ 0.0015  \\
SGD & 0.3285 $\pm$ 0.0139  & 1.7308 $\pm$ 0.0137  & 93.17 $\pm$ 0.14  & 73.15 $\pm$ 0.11  & 0.0483 $\pm$ 0.0022  & 0.1870 $\pm$ 0.0014  \\
SWA & 0.2621 $\pm$ 0.0104  & 1.2780 $\pm$ 0.0051  & 93.61 $\pm$ 0.11  & 74.30 $\pm$ 0.22  & 0.0408 $\pm$ 0.0019  & 0.1514 $\pm$ 0.0032  \\
SGLD & 0.2001 $\pm$ 0.0059  & 0.9699 $\pm$ 0.0057  & 93.55 $\pm$ 0.15  & 74.02 $\pm$ 0.30  & \textbf{0.0082 $\pm$ 0.0012}  & 0.0424 $\pm$ 0.0029  \\
KFAC & 0.2252 $\pm$ 0.0032  & 1.1915 $\pm$ 0.0199  & 92.65 $\pm$ 0.20  & 72.38 $\pm$ 0.23  & 0.0094 $\pm$ 0.0005  & 0.0778 $\pm$ 0.0054  \\
\midrule[0.3pt]\bottomrule[1pt]
\end{tabular}
}
\end{adjustbox}
\label{tab: accuracy for classification}
\end{table*}

\textbf{Results.}
We present the test log likelihoods for small UCI datasets in Table~\ref{tab: small likelihood} and those for large UCI datasets in Table~\ref{tab: large likelihood}.
In both tables,
the first block summarizes SGD-trajectory sampling-based approaches and the second summarizes the rest.
Underlined results are the best among all and bold results are the best among SGD-trajectory sampling-based approaches.
From the results,
our \ciber has substantially better performance than all others on three out of the five small UCI datasets
four out of six large UCI datasets,
with comparable performance on the rest,
demonstrating that \ciber provides accurate uncertainty estimation.
We also present the test rooted-mean-squared error results in Appendix,
where \ciber outperforms all other SGD-trajectory sampling-based baselines on four out of five small UCI datasets and four out of six large UCI datasets; it outperforms all baselines on two small UCI datasets and one large UCI datasets and has comparable performance on the rest.
This 
further illustrates that
exact marginalization over conditional approximate posteriors enabled by WMI solvers achieves accurate estimation of the true BMA and boosts predictive performance.

\begin{table*}[t]
\centering
\caption{Average test performance for image transfer learning tasks.}
\begin{adjustbox}{max width=\textwidth}
\sc
{\setlength\doublerulesep{1pt} 
\begin{tabular}{@{}lllllll@{}}
\toprule[1pt]\midrule[0.3pt]
Metric & \multicolumn{2}{c}{NLL} & \multicolumn{2}{c}{ACC} & \multicolumn{2}{c}{ECE} \\
\midrule
Model & VGG-16 & PreResNet-164 & VGG-16 & PreResNet-164 & VGG-16 & PreResNet-164 \\
\midrule
CIBER & \textbf{0.9869 $\pm$ 0.0102}  & \textbf{0.9684 $\pm$ 0.0075} & \textbf{72.56 $\pm$ 0.23}  & 75.70 $\pm$ 0.17 & \textbf{0.0925 $\pm$ 0.0028}  & \textbf{0.0704 $\pm$ 0.0031}\\
SWAG & 1.3425 $\pm$ 0.0015  & 1.3842 $\pm$ 0.0122  & 72.30 $\pm$ 0.11  & \textbf{76.30 $\pm$ 0.06} & 0.1988 $\pm$ 0.0028  & 0.1668 $\pm$ 0.0006 \\
SGD & 1.6528 $\pm$ 0.0390  & 1.4790 $\pm$ 0.0000  & 72.42 $\pm$ 0.07  & 75.56 $\pm$ 0.00 & 0.2149 $\pm$ 0.0027  & 0.1758 $\pm$ 0.0000 \\
SWA & 1.3993 $\pm$ 0.0502  & 1.3552 $\pm$ 0.0000  & 71.92 $\pm$ 0.01  & 76.02 $\pm$ 0.00 & 0.2082 $\pm$ 0.0056  & 0.1739 $\pm$ 0.0000 \\
\midrule[0.3pt]\bottomrule[1pt]
\end{tabular}
}
\end{adjustbox}
\label{tab: accuracy for transfer learning}
\end{table*}

\subsection{Image Classification}
\textbf{CIFAR datasets.} We experiment with two image datasets: CIFAR-10 and CIFAR-100~\citep{krizhevsky2009learning} and evaluate the test performance using three metrics: 1) negative log likelihood~(NLL) that reflects the quality of both uncertainty estimation and prediction accuracy, 2) classification accuracy~(ACC), and 3) expected calibration errors~(ECE)~\citep{naeini2015obtaining} that show the difference between predictive confidence and accuracy and should be close to zero for a well-calibrated approach.

We run \ciber by choosing the collapsed parameter set to be $10$ weights and $100$ weights at the last layer of the neural network models for CIFAR-10 and CIFAR-100 respectively. The weights are chosen using the same heuristic as the one for regression tasks, i.e., to choose the weights whose samples from the SGD trajectories have large variances.
We compare \ciber with strong baselines including SWAG~\citep{maddox2019simple} reproduced by their open-source implementation, standard SGD, SWA~\citep{izmailov2018averaging}, SGLD~\citep{welling2011bayesian} and KFAC~\citep{ritter2018scalable}.

\textbf{Transfer from CIFAR-10 to STL-10.} We further consider a transfer learning task using the model trained on CIFAR-10 to be evaluated on dataset STL-10~\citep{coates2011analysis}.
STL-10 shares nine out of ten classes with the CIFAR-10 dataset but has a different image distribution. 
It is a common benchmark in transfer learning to adapt models trained on CIFAR-10 to STL-10.

\textbf{Results.} We present the test classification performance on dataset CIFAR-10 and CIFAR-100 in Table~\ref{tab: accuracy for classification} and that of transfer learning in Table~\ref{tab: accuracy for transfer learning}. 
The neural network models used in the classification task are VGG-16 networks
and the models used in the transfer learning task are VGG-16 and PreResNet-$164$.
More results using different network architectures are presented in the Appendix. With the same number of samples as SWAG, \ciber outperforms SWAG and other baselines in most evaluations and delivers comparable performance otherwise, demonstrating the effectiveness of using collapsed samples in improving uncertainty estimation as well as classification performance.

\section{Conclusions And Future Work}
We reveal the connection between BMA, a way to perform Bayesian deep learning and WVC,
which inspires us to approximate BMA using the framework of WMI.
To further make this approximation scalable and flexible, we combine it with collapsed samples which gives our algorithm \ciber.
\ciber compares favorably to Bayesian deep learning baselines on regression and classification tasks.
A future direction would be to explore what other layers can be expressed as SMT formulas and thus amenable to SMT encoding.
Also, the current WMI solvers are limited to polynomial weights, and thus the reduction to WMI problems is applicable to piecewise polynomial weights. This limitation might be alleviated in the future by the development of new WMI solvers that allow various weight function families.

\section*{Acknowledgments}
We would like to thank Yuhang Fan and Kareem Ahmed for helpful discussions.
We would also like to thank Wesley Maddox for answering queries on the implementation of baseline SWAG.
This work was funded in part by the DARPA PTG Program under award HR00112220005, the DARPA ANSR program under award FA8750-23-2-0004, 
NSF grants \#IIS-1943641, \#IIS-1956441, \#CCF-1837129, and a gift from RelationalAI. GVdB discloses a financial interest in RelationalAI.
ZZ is supported by an Amazon Doctoral Student Fellowship.

\bibliography{references}
\bibliographystyle{icml2023}

\clearpage
\appendix
\section{Proofs}
\textbf{Proposition~\ref{prop: collapsed}}
\textit{
Let the SMT formula $\theory = \theory_{\relu} \land \theory_{\posterior} \land \theory_{\prediction}$, and the set of weights $\Weights = \Weights_{\posterior} \cup \Weights_{\prediction}$ as defined in Section~\ref{sec: probs smt encoding}. Let the set of weights
$\Weights^* = \Weights \cup \{\weight_{\literal}(Y) = Y ~~\textit{with}~~ \literal = \true \}$.
The integrals in collapsed BMA (Equation~\ref{eq: collapsed BMA}) can be computed by WMI solvers as
\begin{align*}
    \int p(\y \mid \x, \fw)~q(\fw_{\collapse}) \,d\fw_{\collapse} 
    &=
    \WMI(\theory \land (\bm{Y} = \y), \Weights)
    ~/~
    \WMI(\theory, \Weights)
, \text{ and}\\
    \int \y~p(\y \mid \x, \fw)~~q(\fw_{\collapse}) \,d\fw_{\collapse}~d\y
    &=
        \WMI(\theory, \Weights^*)
    ~/~
        \WMI(\theory, \Weights).
\end{align*}
}
\begin{proof}
By construction, it holds that
\begin{align*}
    p(\y \mid \x, \fw) 
    &\propto
    \WMI(\theory_{\prediction} \land (\Y = \y), \Weights_{\prediction}) \\
    &\propto
    \prod_{\literal \in \literals_{\prediction}} \weight_{\literal}(\y, \fw_{\collapse})^{\id{\y, \fw_{\collapse} \models \literal}},
    ~~\textit{with}~~ (\y, \fw_{\collapse}) \models \theory_{\prediction} \land \theory_{\relu} \\
    q(\fw_{\collapse}) 
    &\propto
    \WMI(\theory_{\posterior} \land (\W_{\collapse} = \fw_{\collapse}), \Weights_{\posterior}) \\
    &\propto
    \prod_{\literal \in \literals_{\posterior}} \weight_{\literal}(\fw_{\collapse})^{\id{\fw_{\collapse} \models \literal}},
    ~~\textit{with}~~ \fw_{\collapse} \models \theory_{\posterior} \\
\end{align*}
Thus, we have that the likelihood weighted by the approximate posterior would be
\begin{align*}
    p(\y \mid \x, \fw)~q(\fw_{\collapse}) 
    &\propto \prod_{\literal \in \literals_{\prediction} \land \literals_{\posterior}} \weight_{\literal}(\y, \fw_{\collapse})^{\id{\y, \fw_{\collapse} \models \literal}}
    ~~\textit{with}~~  (\y, \fw_{\collapse}) \models \theory_{\relu} \land  \theory_{\prediction} \land \theory_{\posterior},
\end{align*}
or equivalently, 
\begin{align*}
    p(\y \mid \x, \fw)~q(\fw_{\collapse})
    &= \frac
    {\prod_{\literal \in \literals_{\prediction} \land \literals_{\posterior}} \weight_{\literal}(\y, \fw_{\collapse})^{\id{\y, \fw_{\collapse} \models \literal}}}
    {\WMI(\theory, \Weights)},
    ~~\textit{with}~~  (\y, \fw_{\collapse}) \models \theory.
\end{align*}
By integrating over the collapsed set $\W_{\collapse}$, it further holds that
\begin{align*}
    &\int p(\y \mid \x, \fw)~q(\fw_{\collapse}) \,d \fw_{\collapse} \\
    &= \frac
    {\int \prod_{\literal \in \literals_{\prediction} \land \literals_{\posterior}} \weight_{\literal}(\y, \fw_{\collapse})^{\id{\y, \fw_{\collapse} \models \literal}} \,d \fw_{\collapse}}
    {\WMI(\theory, \Weights)},
    ~~\textit{with}~~ (\y, \fw_{\collapse}) \models \theory \\
    &= \frac{\WMI(\theory \land (\bm{Y} = \y), \Weights)}
    {\WMI(\theory, \Weights)}
\end{align*}
which proves the first equation.

Similarly, we have that
\begin{align*}
    &\y~p(\y \mid \x, \fw)~q(\fw_{\collapse}) \\
    &\propto \prod_{\literal \in \literals_{\prediction}} \y~\weight_{\literal}(\y, \fw_{\collapse})^{\id{\y, \fw_{\collapse} \models \literal}}
    \prod_{\literal \in \literals_{\posterior}} \weight_{\literal}(\fw_{\collapse})^{\id{\fw_{\collapse} \models \literal}},
    ~~\textit{with}~~(\y, \fw_{\collapse}) \models \theory
\end{align*}
By integrating over the collapsed set $\W_{\collapse}$ and prediction $\y$, it holds that
\begin{align*}
    &\int \y~p(\y \mid \x, \fw)~q(\fw_{\collapse}) \,d \fw_{\collapse}\,d \y\\
    &= \frac
    {\int 
    y \prod\limits_{\literal \in \literals_{\prediction}}
    \weight_{\literal}(\y, \fw_{\collapse})^{\id{\y, \fw_{\collapse} \models \literal}}
    \prod\limits_{\literal \in \literals_{\posterior}} \weight_{\literal}(\fw_{\collapse})^{\id{\fw_{\collapse} \models \literal}} \,d \fw_{\collapse} \,d \y}
    {\WMI(\theory, \Weights)},
    ~~\textit{with}~~ (\y, \fw_{\collapse}) \models \theory \\
    &= \frac{
        \WMI(\theory, \Weights^*)
    }{
        \WMI(\theory, \Weights)
    }
\end{align*}
which finishes our proof.
\end{proof}

\section{Pseudo Code for \ciber}

We summarize our proposed algorithm \textbf{\ciber}, \textbf{C}ollapsed \textbf{I}nference \textbf{B}ayesian D\textbf{E}ep Lea\textbf{R}ning, for regression tasks, in Algorithm~\ref{algo: ciber}.
For the classification task, the algorithm is basically the same except the encoding of the predictive of the distribution.
Specifically, for a given class $y$,
the predictive distribution $p(\y \mid \x, \fw)$ can be encoded into a WMI problem as shown below:
\begin{align*}
\boxed{
\resizebox{.99\textwidth}{!}{$
\begin{array}{l}
\theory_{\prediction} = 
\begin{array}{l}
    \nn_{\fw}(\x) \geq - d
\end{array}
~~
\Weights_{\prediction} =
\left\{
\begin{array}{ll}
    \weight_{\literal_1}(\W_{\collapse})
    &~~\textit{with}~~ \ell_1 = (\nn_{\fw}(\x) \leq d)\\
    \weight_{\literal_2}(\W_{\collapse}) = 1
    &~~\textit{with}~~ \ell_2 = (\nn_{\fw}(\x) > d)\\
\end{array}
\right\}
\end{array}
$}
}
\end{align*}
where $\weight_{\literal_1}$ is a cubic polynomial that approximates the sigmoid function such that the posterior predictive distribution $p(\y \mid \x)$ can be solved by WMI solvers by $p(\y \mid \x) = \WMI(\theory, \Weights)$.
Further, the prediction of BMA for classification tasks is made by $y^* = \argmax_{\y} p(\y \mid \x)$.

\begin{algorithm}[t]
\textbf{Input}:
input $\x$,
sampled weights $\sampleW$, neural network model $\nn_{\fw}$, 
prediction ground truth $y^*$\\
\textbf{Ouput}:
predictions and likelihoods
\begin{algorithmic}[1]
    \State Choose a partition $(\W_{\sampling}, \W_{\collapse})$ for network parameters
    \State 
    Derive approximate posterior $q(\fw_{\collapse})$ from sampled weights $\{\fw_{\collapse} \mid \fw \in \sampleW\}$
    \Comment{cf. Section~\ref{sec: probs smt encoding}}
    \State Encode posterior 
    $q(\fw_{\collapse})$
    into WMI problem
    $\probsmt_{\posterior} = (\theory_{\posterior}, \Weights_{\posterior})$
    \Comment{cf. Section~\ref{sec: probs smt encoding}}
    \State $\mathcal{Y} \leftarrow \emptyset$, $\mathcal{P} \leftarrow \emptyset$
    \Comment{Initialization}
    \For{sample $\fw_{\sampling}$ in $\{\fw_{\sampling} \mid \fw \in \sampleW\}$ }
    \State Encode neural network model $\nn_{\relu}$ parameterized by $(\fw_{\sampling}, \W_{\collapse})$ into an SMT formula $\theory_{\nn_{\fw}}$
    \State Encode predictive $p(Y \mid \x, \fw_{\sampling}, \W_{\collapse})$
    into a WMI problem $\probsmt_{\prediction} = (\theory_{\prediction}, \Weights_{\prediction})$
    \State SMT formula $\theory \leftarrow \theory_{\relu} \land \theory_{\posterior} \land \theory_{\prediction}$
    \State
    Weights $\Weights \leftarrow \Weights_{\posterior} \cup \Weights_{\prediction}$
    \State Weights
    $\Weights^* \leftarrow \Weights \cup \{\weight_{\literal}(Y) = Y ~~\textit{with}~~ \literal = \true \}$
    \State Add prediction $y = \WMI(\theory, \Weights^*) / \WMI(\theory, \Weights)$ to prediction set $\mathcal{Y}$
    \Comment{cf. Section~\ref{sec: Exact Integration in Collapsed BMA}}
    \State Add likelihood $p =
    \WMI(\theory \land (Y = y^*), \Weights) / \WMI(\theory, \Weights)$ to set $\mathcal{P}$
    \Comment{cf. Section~\ref{sec: Exact Integration in Collapsed BMA}}
    \EndFor
    \State \textbf{return} 
    $y = \textsc{Mean}(\mathcal{Y})$, $p(y^* \mid \x) = \textsc{Mean}(\mathcal{P})$
\end{algorithmic}
\caption{CIBER}
\label{algo: ciber}
\end{algorithm}

\begin{table*}[t]
\centering
\caption{Average test RMSE for the small UCI regression task.}
\begin{adjustbox}{max width=\textwidth}
\sc
{\setlength\doublerulesep{1pt} 
\begin{tabular}{@{}llllll@{}}
\toprule[1pt]\midrule[0.3pt]
& Boston & Concrete  & Yacht   & Naval  & Energy \\
\midrule
\ciber(second) & 3.488 $\pm$ 1.123 & 4.880 $\pm$ 0.506 & 0.828 $\pm$ 0.241 & \underline{\textbf{0.000 $\pm$ 0.000}} & \underline{\textbf{0.447 $\pm$ 0.081}} \\
\ciber(last) & 3.478 $\pm$ 1.128 & \textbf{4.854 $\pm$ 0.503} & \textbf{0.752 $\pm$ 0.294} & \underline{\textbf{0.000 $\pm$ 0.000}} & \underline{\textbf{0.447 $\pm$ 0.081}} \\
SWAG & 3.517 $\pm$ 0.981 & 5.233 $\pm$ 0.417 & 0.973 $\pm$ 0.375 & 0.001 $\pm$ 0.000 & 1.594 $\pm$ 0.273 \\
PCA+ESS (SI) & 3.453 $\pm$ 0.953 & 5.194 $\pm$ 0.448 & 0.972 $\pm$ 0.375 & 0.001 $\pm$ 0.000 & 1.598 $\pm$ 0.274 \\
PCA+VI (SI) & \textbf{3.457 $\pm$ 0.951} & 5.142 $\pm$ 0.418 & 0.973 $\pm$ 0.375 & 0.001 $\pm$ 0.000 & 1.587 $\pm$ 0.272 \\
\hline
SGD & 3.504 $\pm$ 0.975 & 5.194 $\pm$ 0.446 & 0.973 $\pm$ 0.374 & 0.001 $\pm$ 0.000 & 1.602 $\pm$ 0.275 \\
MCD & 2.830 $\pm$ 0.170 & 4.930 $\pm$ 0.140 & 0.720 $\pm$ 0.050 & \underline{0.000 $\pm$ 0.000} & 1.080 $\pm$ 0.030 \\
VSD & \underline{2.640 $\pm$ 0.170} & \underline{4.720 $\pm$ 0.110} & \underline{0.690 $\pm$ 0.060} & \underline{0.000 $\pm$ 0.000} & 0.470 $\pm$ 0.010 \\
\midrule[0.3pt]\bottomrule[1pt]
\end{tabular}
}
\end{adjustbox}
\label{tab: small rmse}
\end{table*}

\begin{table*}[t]
\centering
\caption{Average test RMSE for the large UCI regression task.}
\begin{adjustbox}{max width=\textwidth}
\sc
{\setlength\doublerulesep{1pt} 
\begin{tabular}{@{}lllllll@{}}
\toprule[1pt]\midrule[0.3pt]
& Elevators & KeggD  & KeggU   & Protein  & Skillcraft & Pol \\
\midrule
\ciber(second) & \textbf{0.088 $\pm$ 0.002} & 0.142 $\pm$ 0.074 & \textbf{0.115 $\pm$ 0.007} & 0.438 $\pm$ 0.009 & \textbf{0.251 $\pm$ 0.010} & 2.212 $\pm$ 0.230 \\
\ciber(last) & \textbf{0.088 $\pm$ 0.002} & 0.142 $\pm$ 0.072 & 0.118 $\pm$ 0.012 & 0.438 $\pm$ 0.009 & \textbf{0.251 $\pm$ 0.010} & \underline{\textbf{2.199 $\pm$ 0.182}} \\
SWAG & \textbf{0.088 $\pm$ 0.001} & 0.129 $\pm$ 0.029 & 0.160 $\pm$ 0.043 & \underline{\textbf{0.415 $\pm$ 0.018}} & 0.293 $\pm$ 0.015 & 3.110 $\pm$ 0.070 \\
PCA+ESS (SI) & 0.089 $\pm$ 0.002 & 0.129 $\pm$ 0.028 & 0.160 $\pm$ 0.043 & 0.425 $\pm$ 0.017 & 0.293 $\pm$ 0.015 & 3.755 $\pm$ 6.107 \\
PCA+VI (SI) & \textbf{0.088 $\pm$ 0.001} & \textbf{0.128 $\pm$ 0.028} & 0.160 $\pm$ 0.043 & 0.418 $\pm$ 0.021 & 0.293 $\pm$ 0.015 & 2.499 $\pm$ 0.684 \\
\hline
SGD & 0.103 $\pm$ 0.035 & 0.132 $\pm$ 0.017 & 0.186 $\pm$ 0.034 & 0.436 $\pm$ 0.011 & 0.288 $\pm$ 0.014 & 3.900 $\pm$ 6.003 \\
NL & 0.101 $\pm$ 0.002 & 0.134 $\pm$ 0.036 & 0.120 $\pm$ 0.003 & 0.447 $\pm$ 0.012 & 0.253 $\pm$ 0.011 & 4.380 $\pm$ 0.853 \\
DKL & \underline{0.084 $\pm$ 0.020} & \underline{0.100 $\pm$ 0.010} & \underline{0.110 $\pm$ 0.000} & 0.460 $\pm$ 0.010 & \underline{0.250 $\pm$ 0.000} & 6.617 \\
OrthVGP & 0.095 & 0.120 & 0.117 & 0.461 & --- & 4.300 $\pm$ 0.200 \\
FF & 0.089 $\pm$ 0.002 & 0.120 $\pm$ 0.000 & 0.120 $\pm$ 0.000 & 0.470 $\pm$ 0.010 & \underline{0.250 $\pm$ 0.020} & --- \\
\midrule[0.3pt]\bottomrule[1pt]
\end{tabular}
}
\end{adjustbox}
\label{tab: large rmse}
\end{table*}

\begin{table*}[t]
\centering
\caption{Average test log likelihoods for image classification tasks on CIFAR-10 and CIFAR-100. }
\begin{adjustbox}{max width=\textwidth}
\sc
{\setlength\doublerulesep{1pt} 
\begin{tabular}{@{}lllllll@{}}
\toprule[1pt]\midrule[0.3pt]
& \multicolumn{3}{c}{CIFAR-10} & \multicolumn{3}{c}{CIFAR-100} \\
\midrule
Model & VGG-16 & PreResNet-164 & WideResNet & VGG-16 & PreResNet-164 & WideResNet \\
\midrule
CIBER & \textbf{0.1927 $\pm$ 0.0029}  & \textbf{0.1352 $\pm$ 0.0014}  & 0.1913 $\pm$ 0.0029  & \textbf{0.9193 $\pm$ 0.0027}  & 0.8144 $\pm$ 0.0065  & 0.7930 $\pm$ 0.0065  \\
SWAG & 0.2503 $\pm$ 0.0081  & 0.1459 $\pm$ 0.0013  & 0.1076 $\pm$ 0.0009  & 1.2785 $\pm$ 0.0031  & 1.0703 $\pm$ 0.4861  & 0.6719 $\pm$ 0.0035  \\
SGD & 0.3285 $\pm$ 0.0139  & 0.1814 $\pm$ 0.0025  & 0.1294 $\pm$ 0.0022  & 1.7308 $\pm$ 0.0137  & 0.9465 $\pm$ 0.0191  & 0.7958 $\pm$ 0.0089  \\
SWA & 0.2621 $\pm$ 0.0104  & 0.1450 $\pm$ 0.0042  & \textbf{0.1075 $\pm$ 0.0004}  & 1.2780 $\pm$ 0.0051  & 0.7370 $\pm$ 0.0265  & \textbf{0.6684 $\pm$ 0.0034}  \\
SGLD & 0.2001 $\pm$ 0.0059  & 0.1418 $\pm$ 0.0005  & 0.1289 $\pm$ 0.0009  & 0.9699 $\pm$ 0.0057  & \textbf{0.6981 $\pm$ 0.0052}  & 0.6780 $\pm$ 0.0022  \\
KFAC & 0.2252 $\pm$ 0.0032  & 0.1471 $\pm$ 0.0012  & 0.1210 $\pm$ 0.0020  & 1.1915 $\pm$ 0.0199  & 0.7881 $\pm$ 0.0025  & 0.7692 $\pm$ 0.0092  \\
\midrule[0.3pt]\bottomrule[1pt]
\end{tabular}
}
\end{adjustbox}
\label{tab: nll for classification}
\end{table*}

\section{Additional Experiments}

\subsection{Toy Regression in Figure~\ref{fig:toy}}
We evaluate the predictive distributions obtained by our \ciber and HMC respectively, in a toy dataset generated by sampling $10$ input $x$ uniformly distributed in the interval $[-1, -0.5]$ and interval $[0.5, 1]$. 
For each input $x$, the corresponding target $y$ is computed from a cubic polynomial with Gaussian noises. We apply to these data a Bayesian neural network which is a \relu neural network with two hidden layers, where both parameter priors and likelihood are Gaussian distributions. We compare HMC and our \ciber in a few-sample setting which is common in most Bayesian deep learning applications, with $10$ samples from the posterior distribution. 
An estimation generated by HMC with a sufficiently large number of samples of size $2,000$ is further presented as a ground truth.

The results are shown in Figure~\ref{fig:toy}. Even with the same $10$ samples drawn from the posterior distribution, since \ciber further approximates the $10$ samples with a uniform distribution as $q(\fw)$, it yields a predictive distribution $p(\y \mid \x)$ closer to the ground truth than HMC. The intuition behind is that using a uniform distribution instead of a few samples forms a better approximation to the true posterior since the uniform distribution in a collapsed sample represents uncountably many models.

\subsection{Regression on Small and Large Datasets}

\textbf{Sampling from SGD Trajectories.\ }
During training, we use Gaussian log likelihood as the objective for obtaining smooth gradients and use early stopping to prevent over-fitting. At convergence, we start the sampling process by keeping running SGD and collecting the weights. 
At deployment time, we approximate the Gaussian predictive distribution with the triangular distributions.

\textbf{Hyperparameters.\ }
The hyperparameters including learning rates and weight decay are tuned by performing a grid search to maximize the Gaussian log likelihood using a validation split.

\textbf{Additional Results.\ }
Following the set-up of \citet{izmailov2020subspace},
the test performance on small UCI datasets are averaged over $20$ trials and the test performance of large UCI datasets are averaged over $10$ trials.
The test log likelihood results for small UCI datasets are unnormalized while those for large UCI datasets are normalized.
Besides the log likelihood results as shown in Section~\ref{sec: Regression on Small and Large UCI Datasets},
root-mean-squared-error~(RMSE) results for small UCI datasets are further presented in Table~\ref{tab: small rmse} and RMSE results for large UCI datasets are presented in Table~\ref{tab: large rmse}.

\subsection{Image Classification}

\textbf{Sampling from SGD Trajectories.\ }
All the network models are trained for $300$ epochs using SGD.
We start the weight collection after epoch $160$ with step size $5$. We follow exactly the same hyperparameters as \citet{maddox2019simple} including learning rates and weight decay parameters.

\textbf{Additional Results.\ }
Following the set-up of \citet{maddox2019simple},
we run experiments with VGG-16, PreResNet-164 and WideResNet network models on both the image classification task and the transfer learning task.
For the image classification task on CIFAR datasets,
we present the log likelihood results in Table~\ref{tab: nll for classification}, the accuracy results in Table~\ref{tab: accuracy for classification appendix}, and ECE results in Table~\ref{tab: ECE for classification}.
For the transfer learning task from dataset CIFAR-10 to dataset STL-10,
we present the results for model VGG-$16$ and PreResNet-$164$ in Table~\ref{tab: accuracy for transfer learning} and the results for model WideResNet in Table~\ref{tab: test performance wideresnet}.

\begin{table*}[t]
\centering
\caption{Average test accuracy for image classification tasks on CIFAR-10 and CIFAR-100. }
\begin{adjustbox}{max width=\textwidth}
\sc
{\setlength\doublerulesep{1pt} 
\begin{tabular}{@{}lllllll@{}}
\toprule[1pt]\midrule[0.3pt]
& \multicolumn{3}{c}{CIFAR-10} & \multicolumn{3}{c}{CIFAR-100} \\
\midrule
Model & VGG-16 & PreResNet-164 & WideResNet & VGG-16 & PreResNet-164 & WideResNet \\
\midrule
CIBER & \textbf{93.64 $\pm$ 0.09}  & 95.95 $\pm$ 0.06  & 95.63 $\pm$ 0.16  & \textbf{74.71 $\pm$ 0.18} & 79.23 $\pm$ 0.25  & 81.25 $\pm$ 0.35  \\
SWAG & 93.59 $\pm$ 0.14  & \textbf{96.09 $\pm$ 0.08}  & 96.38 $\pm$ 0.08  & 73.85 $\pm$ 0.25  & 73.02 $\pm$ 10.30  & 82.27 $\pm$ 0.07  \\
SGD & 93.17 $\pm$ 0.14  & 95.49 $\pm$ 0.06  & 96.41 $\pm$ 0.10  & 73.15 $\pm$ 0.11  & 78.50 $\pm$ 0.32  & 80.76 $\pm$ 0.29  \\
SWA & 93.61 $\pm$ 0.11  & \textbf{96.09 $\pm$ 0.08}  & \textbf{96.46 $\pm$ 0.04}  & 74.30 $\pm$ 0.22  & \textbf{80.19 $\pm$ 0.52}  & \textbf{82.40 $\pm$ 0.16}  \\
SGLD & 93.55 $\pm$ 0.15  & 95.55 $\pm$ 0.04  & 95.89 $\pm$ 0.02  & 74.02 $\pm$ 0.30  & 80.09 $\pm$ 0.05  & 80.94 $\pm$ 0.17  \\
KFAC & 92.65 $\pm$ 0.20  & 95.49 $\pm$ 0.06  & 96.17 $\pm$ 0.00  & 72.38 $\pm$ 0.23  & 78.51 $\pm$ 0.05  & 80.94 $\pm$ 0.41  \\
\midrule[0.3pt]\bottomrule[1pt]
\end{tabular}
}
\end{adjustbox}
\label{tab: accuracy for classification appendix}
\end{table*}

\begin{table*}[t]
\centering
\caption{Average test ECE for image classification tasks on CIFAR-10 and CIFAR-100. }
\begin{adjustbox}{max width=\textwidth}
\sc
{\setlength\doublerulesep{1pt} 
\begin{tabular}{@{}lllllll@{}}
\toprule[1pt]\midrule[0.3pt]
& \multicolumn{3}{c}{CIFAR-10} & \multicolumn{3}{c}{CIFAR-100} \\
\midrule
Model & VGG-16 & PreResNet-164 & WideResNet & VGG-16 & PreResNet-164 & WideResNet \\
\midrule
CIBER & 0.0130 $\pm$ 0.0011  & 0.0250 $\pm$ 0.0005  & 0.0760 $\pm$ 0.0011  & \textbf{0.0168 $\pm$ 0.0025}  & 0.1423 $\pm$ 0.0029  & 0.1650 $\pm$ 0.0046  \\
SWAG & 0.0391 $\pm$ 0.0020  & 0.0214 $\pm$ 0.0005  & 0.0096 $\pm$ 0.0006  & 0.1535 $\pm$ 0.0015  & 0.1031 $\pm$ 0.0471  & 0.0678 $\pm$ 0.0006  \\
SGD & 0.0483 $\pm$ 0.0022  & 0.0255 $\pm$ 0.0009  & 0.0166 $\pm$ 0.0007  & 0.1870 $\pm$ 0.0014  & 0.1012 $\pm$ 0.0009  & 0.0479 $\pm$ 0.0010  \\
SWA & 0.0408 $\pm$ 0.0019  & 0.0203 $\pm$ 0.0010  & 0.0087 $\pm$ 0.0002  & 0.1514 $\pm$ 0.0032  & 0.0700 $\pm$ 0.0056  & 0.0684 $\pm$ 0.0022  \\
SGLD & \textbf{0.0082 $\pm$ 0.0012}  & 0.0251 $\pm$ 0.0012  & 0.0192 $\pm$ 0.0007  & 0.0424 $\pm$ 0.0029  & 0.0363 $\pm$ 0.0008  & \textbf{0.0296 $\pm$ 0.0008}  \\
KFAC & 0.0094 $\pm$ 0.0005  & \textbf{0.0092 $\pm$ 0.0018}  & \textbf{0.0060 $\pm$ 0.0003}  & 0.0778 $\pm$ 0.0054  & \textbf{0.0158 $\pm$ 0.0014}  & 0.0379 $\pm$ 0.0047  \\
\midrule[0.3pt]\bottomrule[1pt]
\end{tabular}
}
\end{adjustbox}
\label{tab: ECE for classification}
\end{table*}

\begin{table*}[t]
\centering
\caption{Average test performance for image transfer learning tasks using WideResNet.}
\begin{adjustbox}{max width=0.6\textwidth}
\sc
{\setlength\doublerulesep{1pt} 
\begin{tabular}{@{}llll@{}}
\toprule[1pt]\midrule[0.3pt]
Metric & {NLL} & {ACC} & {ECE} \\
\midrule
CIBER & \textbf{0.8259 $\pm$ 0.0148} & 75.02 $\pm$ 0.31 & \textbf{0.0336 $\pm$ 0.0009}\\
SWAG & 1.0142 $\pm$ 0.0032 & 76.96 $\pm$ 0.08  & 0.1303 $\pm$ 0.0008  \\
SGD & 1.1308 $\pm$ 0.0000 & 76.75 $\pm$ 0.00 & 0.1561 $\pm$ 0.0000  \\
SWA & 1.0047 $\pm$ 0.0000 & \textbf{77.50 $\pm$ 0.00} & 0.1413 $\pm$ 0.0000  \\
\midrule[0.3pt]\bottomrule[1pt]
\end{tabular}
}
\end{adjustbox}
\label{tab: test performance wideresnet}
\end{table*}

\end{document}